\documentclass[runningheads]{llncs}
\usepackage{graphicx}

\usepackage{tikz}
\usepackage{comment}
\usepackage{amsmath,amssymb} 
\usepackage{color}

\usepackage[accsupp]{axessibility}  

\usepackage{subcaption,booktabs}
\usepackage{sidecap}
\usepackage{varwidth}
\usepackage{tabularx}
\usepackage{float}

\usepackage{amsmath,amsfonts,bm}

\def\eqref#1{equation~\ref{#1}}

\def\1{\bm{1}}

\newcommand{\test}{\mathcal{D_{\mathrm{test}}}}

\def\va{{\bm{a}}}
\def\vb{{\bm{b}}}

\def\vh{{\bm{h}}}

\def\vl{{\bm{l}}}

\def\vn{{\bm{n}}}

\def\vr{{\bm{r}}}
\def\vs{{\bm{s}}}

\def\vv{{\bm{v}}}

\def\mB{{\bm{B}}}

\def\mG{{\bm{G}}}

\def\mI{{\bm{I}}}

\def\mM{{\bm{M}}}

\def\mS{{\bm{S}}}

\DeclareMathAlphabet{\mathsfit}{\encodingdefault}{\sfdefault}{m}{sl}
\SetMathAlphabet{\mathsfit}{bold}{\encodingdefault}{\sfdefault}{bx}{n}

\newcommand{\R}{\mathbb{R}}

\usepackage{xspace}
\makeatletter
\DeclareRobustCommand\onedot{\futurelet\@let@token\@onedot}
\def\@onedot{\ifx\@let@token.\else.\null\fi\xspace}
\def\eg{\emph{e.g}\onedot} 
\def\ie{\emph{i.e}\onedot}

\def\etal{\emph{et al}\onedot}
\makeatother

\usepackage[normalem]{ulem}
\useunder{\uline}{\ul}{}

\usepackage[capitalize]{cleveref}
\crefname{section}{Sec.}{Secs.}
\Crefname{section}{Section}{Sections}
\Crefname{table}{Table}{Tables}
\crefname{table}{Tab.}{Tabs.}

\begin{document}
\pagestyle{headings}
\mainmatter
\def\ECCVSubNumber{5007}  

\title{Self-calibrating Photometric Stereo by Neural Inverse Rendering}

\titlerunning{Self-calibrating Photometric Stereo by Neural Inverse Rendering}
\author{Junxuan Li\index{Li, Junxuan} 
\and Hongdong Li}
\authorrunning{J. Li and H. Li}
\institute{Australian National University\\
\email{\{junxuan.li,hongdong.li\}@anu.edu.au}}
\maketitle

\begin{abstract}
This paper tackles the task of uncalibrated photometric stereo for 3D object reconstruction, where both the object shape, object reflectance, and lighting directions are unknown. This is an extremely difficult task, and the challenge is further compounded with the existence of the well-known generalized bas-relief (GBR) ambiguity in photometric stereo.  Previous methods to resolve this ambiguity either rely on an overly simplified reflectance model, or assume special light distribution. We propose a new method that jointly optimizes object shape, light directions, and light intensities, all under general surfaces and lights assumptions. The specularities are used explicitly to solve uncalibrated photometric stereo via a neural inverse rendering process. We gradually fit specularities from shiny to rough using novel progressive specular bases. Our method leverages a physically based rendering equation by minimizing the reconstruction error on a per-object-basis. Our method demonstrates state-of-the-art accuracy in light estimation and shape recovery on real-world datasets.
\keywords{Uncalibrated photometric stereo; generalized bas-relief ambiguity; neural network; inverse rendering.}
\end{abstract}

\section{Introduction}
Photometric Stereo (PS) aims to reconstruct the 3D shape of an object given a set of images taken under different lights.  Calibrated photometric stereo methods assume the light directions are known in all images~\cite{woodham1980photometric,wu2010robust,ikehata2012robust,mukaigawa2007analysis,wu2010photometric,santo2017deep,ikehata2018cnn,chen2020deep}. However, it is quite a tedious and laborious effort to calibrate the light sources in all input images in practice, often requiring instrumented imaging environment and expert knowledge. How to solve uncalibrated photometric stereo is therefore a crucial milestone to bring PS to practical use. 

Recovering the surface shape with unknown light sources and general reflectance is difficult. Previous methods tackle this problem by assuming the Lambertian surfaces. However, Lambertian surfaces in uncalibrated photometric stereo have an inherent $3\times 3$ parameters ambiguity in normals and light directions~\cite{belhumeur1999bas}. When the surface integrability constraint is introduced, this ambiguity can be further reduced to a 3-parameter generalized bas-relief  (GBR) ambiguity. Additional information is required to further resolve this ambiguity.

Existing methods to resolve the GBR ambiguity resort to introducing additional knowledge, such as priors on the albedo distribution~\cite{alldrin2007resolving}, color intensity profiles~\cite{shi2010self,lu2013uncalibrated}, and symmetric BRDFs~\cite{wu2013calibrating,lu2015uncalibrated,lu2017symps}. 
Drbohlav~\etal~\cite{drbohlav2002specularities} leveraged the mirror-like specularities on a surface to resolve the ambiguity. But they need to manually label the mirror-like specularities for computation.  Georghiades~\cite{georghiades2003incorporating} addressed the ambiguity by using the TS reflectance model~\cite{torrance1967theory}. However, to avoid the local minima, they further assumed the uniformly distributed albedos.
These methods either rely on unrealistic assumptions or are unstable to solve.  Hence, there is still a gap in applying this technique to more generalized real-world datasets. 
Recent deep learning-based methods push the boundary of light estimation and surface normal estimation~\cite{chen2019self,kaya2021uncalibrated}. These methods treated light estimation as a classification task. Hence, they lose the ability to continuously represent the lights.

In this paper, we present an inverse rendering approach for uncalibrated photometric stereo. We propose a model which explicitly uses specular effects on the object's surface to estimate both the lights and surface normals. We show that by incorporating our model, the GBR ambiguity can be resolved up to a binary convex/concave ambiguity. Our neural network is optimized via the inverse rendering error. Hence, there is no need to manually label the specular effects during the process. To avoid local minima during the optimization, we propose \emph{progressive specular bases} to fit the specularities from shiny to rough. The key idea of the above technique is to leverage the mirror-like specularities to reduce GBR ambiguity in the early stage of optimization.  We propose a neural representation to continuously represent the lighting, normal and spatially-varying albedos. By fitting both the specular and diffuse photometric components via the inverse rendering process, our neural network can jointly optimize and refine the estimation of light directions, light intensities, surface normals, and spatially-varying albedos. In summary, our contributions in this paper are:{\small 
\begin{itemize}
\item We propose a neural representation that jointly estimates surface normals, light sources, and albedos via inverse rendering. 
\item We propose progressive specular bases to guide the network during optimization, effectively escaping local minima. 
\end{itemize}} 
Extensive evaluations on challenging real-world datasets show that our method achieves state-of-the-art performance on lighting estimation and shape recovery.

\section{Related Work}


\textbf{Calibrated Photometric Stereo.} By assuming the surface of objects to be ideal Lambertian, shapes can be revealed in closed-form with three or more known lights~\cite{woodham1980photometric}. This restricted assumption is gradually relaxed by following studies~\cite{wu2010robust,mukaigawa2007analysis,wu2010photometric,ikehata2012robust}, where error terms were introduced to account for the deviations from the Lambertian assumption. A regression-based inverse rendering framework~\cite{ikehata2014photometric} was also used for dealing with more general surfaces. Also, in recent years, deep learning-based methods have been widely used in the context of photometric stereo~\cite{santo2017deep,ikehata2018cnn,li2019learning,chen2020deep,yao2020gps,wang2020non,zheng2019spline}. 
Santo~\etal~\cite{santo2017deep} proposed the first photometric stereo neural network, which feeds image pixels into the network in a predetermined order.
Some later works rearranged the pixels into an observation map and then solved the problem per-pixelly~\cite{ikehata2018cnn,li2019learning,zheng2019spline,logothetis2021px}. 
Other deep learning-based approaches used both local and global images cues for normal estimation~\cite{chen2020deep,yao2020gps,wang2020non,honzatko2021leveraging,ju2021recovering}. 
However, their works assumed both the light directions and intensities to be known.
Calibrating light sources may be a tedious process that requires professional knowledge. It will be more convenient to the public if no ground truth light directions are needed for photometric stereo.

\textbf{Uncalibrated Photometric Stereo.} 
Under the Lambertian surface assumption,  there is an inherent generalized bas-relief ambiguity in solving uncalibrated photometric stereo~\cite{belhumeur1999bas}. Traditional works explored many directions to resolve this ambiguity by providing additional knowledge to the system, such as specularities~\cite{drbohlav2002specularities}, TS model~\cite{georghiades2003incorporating}, priors on the albedo distribution~\cite{alldrin2007resolving}, shadows~\cite{sunkavalli2010visibility}, color intensity profiles~\cite{shi2010self,lu2013uncalibrated}, perspective views\cite{papadhimitri2013new}, inter-reflections~\cite{chandraker2005reflections}, local diffuse reflectance maxima~\cite{papadhimitri2014closed}, symmetric BRDFs~\cite{wu2013calibrating,lu2015uncalibrated,lu2017symps}, and total variation~\cite{queau2015solving}. In the presence of inaccurate lighting, Qu\'eau~\etal~\cite{queau2017non} refined the initial lighting estimation by explicitly modeling the outliers among Lambertian assumption. 
Other works aim at solving the uncalibrated photometric stereo under natural illumination~\cite{mo2018uncalibrated,haefner2019variational}; and semi-calibrated lighting where light directions are known but light intensities are unknown~\cite{logothetis2017semi,cho2018semi}.
With the advance of the neural network, deep learning-based methods produced state-of-the-art performance in this area. Chen~\etal~\cite{chen2018ps} proposed a neural network that directly takes images as input, and outputs the surface normal. Later works~\cite{chen2019self,chen2020learned} further improved this pipeline by predicting both the light directions and surface normal at the same time. A recent work~\cite{sarno2021neural} proposes a way to search for the most efficient neural architecture for uncalibrated photometric stereo. These neural network methods learn prior information for solving the GBR ambiguity from a large amount of training data with ground truth. 

\textbf{Neural Inverse Rendering.}
Taniai~\etal~\cite{taniai2018neural} proposed the first neural inverse rendering framework for photometric stereo. They proposed a convolutional neural network that takes images at the input and directly outputs the surface normal. Li~\etal~\cite{li2022neural} proposed an MLP framework for solving the geometry and reflectance via the reconstruction errors. But their works require the light direction at inputs.
Kaya~\etal~\cite{kaya2021uncalibrated} use a pre-trained light estimation network to deal with unknown lights. However, their work cannot propagate the reconstruction error back to the light directions and intensities. 

In this paper, we propose a neural representation that explicitly models the specularities and uses it for resolving the GBR ambiguity via an inverse rendering process. 
Our model allows the re-rendered errors to be back-propagated to the light sources and refines them jointly with the normals. Hence, our method is also robust when accounting for inaccurate lighting.

\section{Specularities Reduce GBR Ambiguity}
In this section, we introduce the notations and formulations of image rendering in the context of uncalibrated photometric stereo under general surfaces. We discuss the GBR ambiguity under Lambertian surfaces. We further demonstrate that the GBR ambiguity can be resolved under non-Lambertian surfaces with the presence of specularities.

\subsection{GBR ambiguity}
Given any point in an object's surface, we assume its surface normal to be $\vn \in \R ^3$. It is illuminated by a distant light with direction to be $\vl \in \R ^3$ and light intensity to be $e \in \R^{+}$. If we observe the surface point from view direction $\vv \in \R^3$, its pixel intensity $m \in \R^{+}$ can be modeled as: $m = e \rho(\vn, \vv, \vl) \max (\vn ^T \vl , 0).$
Here, the $\rho(\vn, \vv, \vl)$ denotes a surface point's BRDF function, which is influenced by the surface normal, view direction, and lighting direction.
The noise, interreflections, and cast-shadows that deviate from the rendering equation are ignored.

In the above equation, traditional methods assume the surface material to be ideal Lambertian, which makes the BRDF function to be a constant: $\rho(\vn, \vv, \vl)=\rho_d \in \R$. For simplicity, we omit the attached-shadows operator $\max (\cdot)$, and incorporate the diffuse albedo and light intensities into the surface normal and light direction.  The equation can be rewrite as 
\begin{align}
    \mM = \mB^T \mS,
\end{align}
where $\mB = [\rho_{d_{1}} \vn_1 , \cdots, \rho_{d_{p}} \vn_p]\in \R ^{3\times p}$ denotes the normal matrix with $p$ different pixels in a image; $\mS = [e_{1} \vl_1 , \cdots, e_n \vl_n]\in \R ^{3\times n}$ denotes the light matrix with $n$ different light sources; $\mM \in \R^{p\times n}$ denotes the $p$ pixels' intensities under $n$ different light sources.
Under this simplified assumption, once the surface point is illuminated by three or more known light sources, the equation has a closed-form solution on surface normals~\cite{woodham1980photometric}. 

Under the uncalibrated photometric stereo setting,  both the light directions and light intensities are unknown. The above equation will have a set of solutions in a $3\times 3$ linear space. By applying the surface integration constraints, it can be further reduced to a $3$ parameters space, which is also known as the generalize bas-relief (GBR) ambiguity in the form as below
\begin{align}\label{eq:g_matrix}
    \mG = \begin{bmatrix}
1 & 0 & 0\\
0 & 1 & 0\\
\mu & \nu & \lambda
\end{bmatrix}
\end{align}
where $\lambda \neq 0; \mu , \nu \in \R$. The transformed normal $\widehat{\mB} = \mG^{-T} \mB$, transformed light $\widehat{\mS} = \mG \mS$.
So that both sides of $\mM = \mB^T \mG^{-1} \mG \mS$  remain equivalent after the transformation. Additional knowledge need to be introduced for solving the GBR ambiguity above.

\begin{SCfigure}[][t]
\caption{Rendered ``Bunny'' and ``Sphere'' using different specular bases with different roughness. The roughness term controls the sharpness of a specular lobe. 
The basis presents narrow specular spikes when roughness is small, which is close to the mirror-like reflection. 
}
\includegraphics[width=0.5\textwidth]{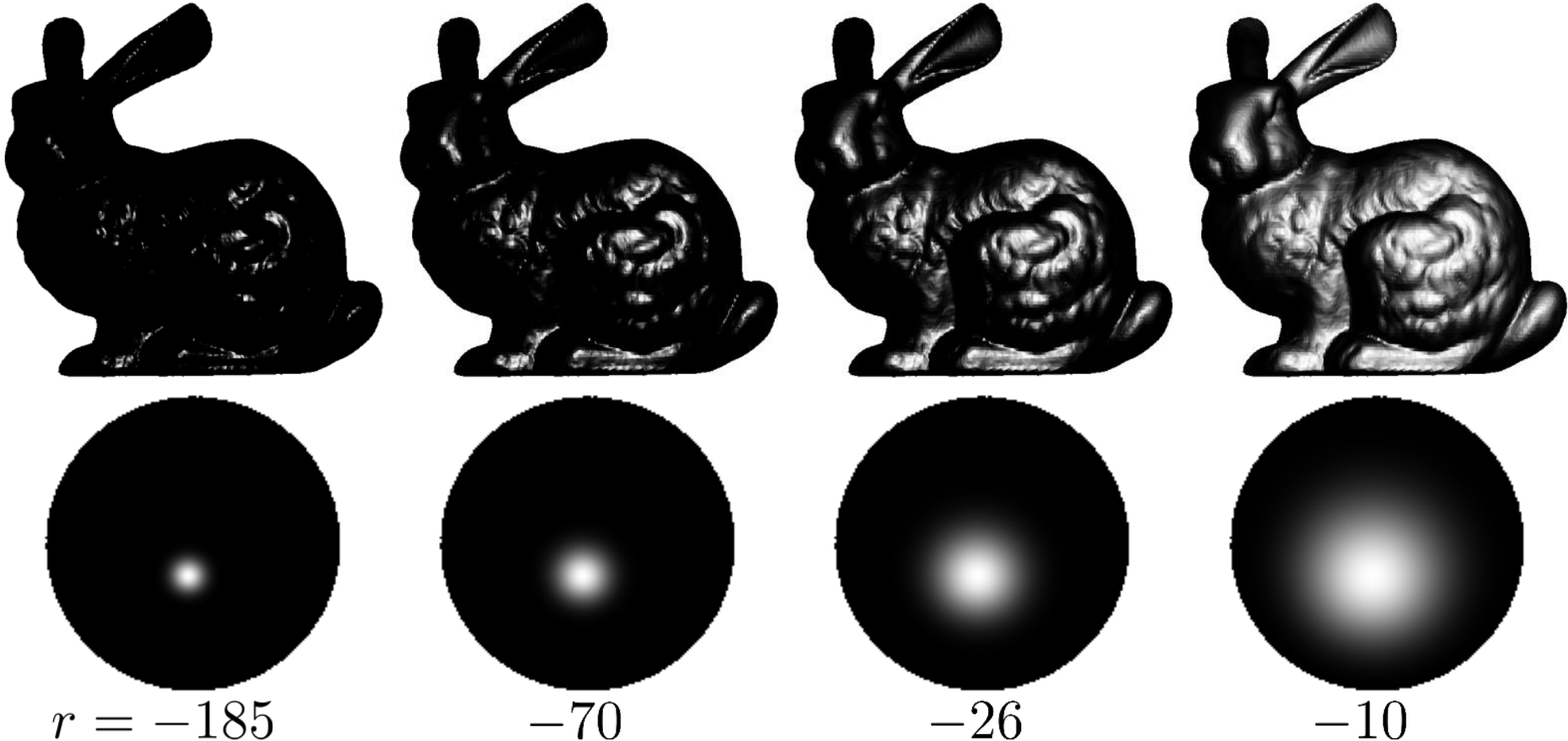}
\label{fig:basis_roughness}
\end{SCfigure}

\subsection{Resolving the ambiguity with specularities}\label{sec:specular_solve_GBR}
We now explain how specularities on object surfaces provide additional information for reducing the GBR ambiguity.
For simplicity, we incorporate the diffuse albedo into surface normal $\vb = \rho_d \vn$, and incorporate the light intensity into light direction $\vs = e \vl$, and only consider the illuminated points. As stated above, the GBR ambiguity exists when we assume the surface to be Lambertian. 
There exists a transformed surface normal and light direction $\widehat{\vb} = \mG^{-T} \vb$, $\widehat{\vs} = \mG \vs$, so that the transformed surface  and lights will compose the identical pixel observation $\widehat{m} = \widehat{\vb}^T \widehat{\vs} = \vb^T \mG^{-1} \mG \vs = m$.

However, in the presence of specularities, the surface BRDF is not constant anymore.
Georghiades~\cite{georghiades2003incorporating} models the reflectance as the combination of the diffuse and specular parts as below\footnote{For simplicity, we omit some terms from \cite{georghiades2003incorporating} without affecting the correctness of their proof.}
\begin{align}\label{eq:brdf_simple}
    \rho (\vn, \vv, \vl) = \rho_d + \rho_{s} \exp(r (1-\vn^T \vh) ), 
\end{align}
where the $\rho_d$ is the diffuse albedo,  $\rho_{s}$ denotes the specular albedo, and the $r\in \R^{-}$ denotes the roughness. $\vh = \frac{\vv + \vl}{||\vv + \vl||}$ is the half-unit-vector between view direction $\vv$ and light direction $\vl$.
Therefore, with the specular terms, the image intensity after the GBR transformation is
\begin{align}
\widehat{m} &= \widehat{\vb}^T \widehat{\vs} +  \rho_{s} ||\widehat{\vs}|| \exp(r (1-  \frac{\widehat{\vb}^T}{||\widehat{\vb}||} \frac{\vv + \frac{\widehat{\vs}}{||\widehat{\vs}||}}{||\vv +\frac{\widehat{\vs}}{||\widehat{\vs}||}||} ) )  \nonumber \\
&= \vb^T \mG^{-1} \mG \vs + 
 \rho_{s}||\mG\vs||
\exp(
r (1-\frac{\vb^T \mG^{-1}} {||\mG^{-T} \vb||} \frac{\vv + \frac{\mG \vs}{||\mG \vs||}}{||\vv + \frac{\mG \vs}{||\mG \vs||}||}
) ),
\end{align}
where $||\cdot||$ denotes the length of a vector. In general, $m = \widehat{m}$ only holds for all pixels in the images when the GBR transformation matrix $\mG$ is identity matrix. 
Theoretical proof was made by Georghiades~\cite{georghiades2003incorporating} that when providing with four different  $(\vb,\vs)$ pairs, it is sufficient to solve the GBR ambiguity up to the binary convex/concave ambiguity, \ie $\lambda=\pm1; \mu,\nu=0$. 
However, even a global minimum  exists on $\mG$, there is no guarantee that no local minima in the 3 parameters space of  $\lambda, \mu$ and $\nu$.  Solving the above equation is still  challenging given the existence of noise, shadows and inter-reflections in real world images. To avoid the local minima,  Georghiades~\cite{georghiades2003incorporating} assumed that the specular albedo $\rho_s$ is uniform across the surface. This uniform specular albedo assumption prevents their method from being applied to more general objects.
In this paper, we are aiming to solve this problem in more general surfaces,  \ie under spatially-varying diffuse and spatially-varying specular albedo.

\paragraph{Resolving the GBR ambiguity by Specular Spikes} 
In fact, the GBR ambiguity can also be resolved by merely four or more pairs of mirror-like reflection effects (\ie specular spikes) on a surface~\cite{drbohlav2002specularities}. The roughness term $r$ in \eqref{eq:brdf_simple} controls the sharpness of a specular lobe. As shown in \cref{fig:basis_roughness}, when roughness is small, the resulted material is very close to a mirror-like material (see $r_i=-185$). 
The specular basis reaches its highest value when $1 - \vn^T \frac{\vv +\vl}{||\vv +\vl||}  =0$. 
Since all the three vectors here are unit vectors, the above equation holds when surface normal $\vn$ is a bisector between the viewing direction $\vv$ and the light direction $\vl$.
Hence, we have the following equation when the basis function reach its highest value, \ie where the mirror-like specularities happens
\begin{align}
    \vv = 2(\vl^T \vn ) \vn - \vl.
\end{align}
From the  \emph{consistent viewpoint constraint}~\cite{drbohlav2002specularities}, the GBR ambiguity can be reduced to two-parameteric group of transformations (rotation around the viewing vector and isotropic scaling). However, the mirror-like specular spikes needed to be manually labeled in previous method~\cite{drbohlav2002specularities}.
While in our paper, these mirror-like specular effects can be automatically fitted via our neural network.

\section{Proposed Method} \label{sec:method}
We propose a neural network based method that aims at inverse rendering the object by factoring the lighting, surface normal,  diffuse albedo, and specular components. 
This section describes our model for solving uncalibrated photometric stereo in the presence of specularities. 

\subsection{Proposed image rendering equation}
Following previous works on uncalibrated photometric stereo, we make the following assumptions on the problem. We assume that the images are taken in orthographic views. Hence the view direction are consistent across the object surface, $\vv = [0,0,1]^T$. The object is only illuminated once by distance lights with unknown direction $\vl$ and intensities $e$.
Given the above assumptions, we now rewrite the rendering equation as
\begin{align}\label{eq:our_image_rendering}
    m = e \rho(\vn, \vl) \max(\vn^T\vl, 0).
\end{align}
Here, the only information we have is the observation of the surface point's pixel intensity $m$. Our target is to inverse this rendering equation to get all the other unknown terms, such as surface normal $\vn$, light direction $\vl$, light intensity $e$, and surface BRDF function $\rho(\cdot)$. In the following sections, we present our model to parameterize and optimize these terms. 

\subsection{BRDF modeling}
As discussed by \eqref{eq:g_matrix} above, the Lambertian surface assumption alone will lead to GBR ambiguity in solving uncalibrated photometric stereo problem.  Hence, we model the reflectance as the combination of the diffuse and specular parts as $\rho (\vn, \vl) = \rho_d +\rho_s(\vn, \vl)$, 
where the $\rho_d$ is the diffuse albedo, and $\rho_s(\vn, \vl)$ is the specular terms. 
We further model the specular term as the summation of a set of specular bases as below
\begin{align}\label{eq:specular_bases_sum}
    \rho_s(\vn, \vl) = \sum_{i=1}^k \rho_{s_i} \exp(r_i (1-\vn^T \vh) ), 
\end{align}
where $\vh = \frac{\vv + \vl}{||\vv + \vl||}$ is the half-unit-vector between view direction $\vv$ and light direction $\vl$; $k$ is the number of bases.
Here, we adopted the Spherical Gaussian~\cite{wang2009all} as our basis function. The $\rho_{s_i}$ denotes the specular albedo, and the $r_i\in \R^{-}$ denotes the roughness. The lower the roughness, the more shiny the material will be. We rendered two objects with the proposed specular basis, as shown in \cref{fig:basis_roughness}.

In summary, our BRDF modeling takes both the diffuse and specular component into consideration and estimate them jointly. 
We also model the specularities as a summation of a set of bases, which enable the material to range from shiny to rough. We can now rewrite the rendering equation as below
\begin{align}
    m = e(\rho_d + \sum_{i=1}^k  \rho_{s_i}\exp(r_i (1-\vn^T \vh))) \max (\vn^T \vl, 0).
\end{align}

\subsection{Progressive Specular Bases}

Inspired by the two ways of resolving GBR in \cref{sec:specular_solve_GBR}, we proposed the novel \emph{progressive specular bases} to solve the uncalibrated photometric stereo robustly.
The key idea of progressive specular bases is to first fit the surface with only  mirror-like specular bases (bases with small roughness term $r_i$); then, we gradually enable the other specular bases for more diffuse effects (bases with large roughness). 

At the early stage of optimization, we only enable mirror-like specular bases, the network will attempt to solve uncalibrated photometric stereo using only the mirror-like specular spikes. Then, as the optimization progresses, other specular bases for the network are gradually enabled to fit those diffuse effects.
Our progressive specular bases will guide the network away from local minima at the early stage of optimization, resulting better optimized results in the end.

The progressive specular bases is achieved by applying a smooth mask on the different specular basis (from small roughness with mirror-like effects to large roughness with less sharp effects) over the course of optimization. 
The weights applied to the different specular bases are defined as below
\begin{align}\label{eq:weighted_specular_bases}
\rho_s(\vn, \vl) = \sum_{i=1}^k \omega_i(\alpha) \rho_{s_i} \exp(r_i (1-\vn^T \vh) ),
\end{align}
where the weight $\omega_i(\alpha)$ is defined as
\begin{align}
      \omega_i(\alpha) =
    \begin{cases}
      0 \quad &\text{if} \ \ \alpha < i \\
      \frac{1-\cos ((\alpha-i)\pi)}{2} \quad &\text{if}\ \  0 \leq \alpha - i < 1 \\
      1 \quad &\text{if} \ \ \alpha - i \geq 1 
    \end{cases}     
\end{align}
$\alpha \in [0, k]$ will gradually increase during the optimization progress. The defined weights above are inspired by a recent coarse-to-fine positional encoding strategy on camera pose estimation~\cite{lin2021barf}.
In the early stage of optimization, the $\alpha$ is small, hence the weight $\omega(\alpha)=0$ will be zero for those specular bases with roughness $r_i$, where $i>\alpha$. 
As the optimization progress, we gradually activate the specular bases one by one. When $\alpha = k$, all the specular bases is used, hence, \eqref{eq:weighted_specular_bases} is identical to \eqref{eq:specular_bases_sum} in the final stage. 
In practice, we set the specular roughness terms $\vr = \{r_i | i \in \{1,\cdots,k\}\}$ in ascending order. So that the above weight will gradually activate the specular bases from small roughness to large roughness. 

To sum up, when applying progressive specular bases, the network will focus on fitting the bright specular spikes at an early stage; then, as the optimization progress, more specular bases are available for the network to fit on the diffuse effects.

\subsection{Neural representation for surfaces}
Here, we describe our neural representation for object surface modeling.
Inspired by the recently proposed coordinately-based multilayer-perceptron (MLP) works~\cite{mildenhall2020nerf}, we proposed two coordinately-based networks which take only the pixel coordinates $(x, y)$ at input, and output the corresponding surface normal and diffuse albedos and specular albedos. 
\begin{align}
    \vn = N_\Theta (x,y),  \\
    \rho_d , \va = M_\Phi (x,y).
\end{align}
Where $N_\Theta(\cdot), M_\Phi(\cdot)$ are MLPs with $\Theta, \Phi$ to be their parameters respectively. Given a image pixel coordinates $(x,y)$, the two MLPs directly output the surface normal $\vn$, diffuse albedo $\rho_d$, and specular albedos $\va = \{\rho_{s_i} | i\in\{1,\cdots,k\}\}$ of that position.

\subsection{Neural representation for lighting}
Next, we describe the parameterization of the light direction and intensity. Let $\mI \in \R^{h\times w}$ denotes the image taken under a light source, where $h,w$ denote the height and width of the input image.  The direction and intensity of that light source are directly predicted by feeding this image into a convolutional neural network:
\begin{align}
e, \vl = L_\Psi (\mI).
\end{align}
where $L_\Psi (\cdot)$ is a convolutional neural network with its parameters $\Psi$. The network $L_\Psi (\cdot)$ takes only the image $\mI$ as input, directly output the corresponding light direction $\vl$ and light intensity $e$. 
Unlike previous deep learning based lighting estimation network~\cite{chen2019self,chen2020learned,kaya2021uncalibrated}, we do not fix the lighting estimation at testing. Instead, the lighting estimation is further refined (\ie fine-tuned) on the testing images by jointly optimizing the lighting, surface normals, and albedos via the reconstruction loss.

\section{Implementation}
This section describes the detail of network architectures, hyperparameters selection, and loss functions.
\paragraph{Network architectures}
The surface normal net $N_\Theta(\cdot)$ uses $8$ fully-connected layers with $256$ channels, followed by a ReLU activation function except for the last layer.
The material net $M_\Phi(\cdot)$ uses the same structure but with $12$ fully-connected ReLU layers.
We apply a positional encoding strategy with $10$ levels of Fourier functions to the input pixel coordinates $(x,y)$ before feeding them to the normal and material MLPs. The lighting network $L_\Psi (\cdot)$ consists of $7$ convolutional ReLU layers and $3$ fully connected layers. Please see the supplementary material for detailed network architectures.

For the choices of specular bases, we initialize the roughness value for each basis range from $-r_{t}$ to  $-r_{b}$ with logarithm intervals
\begin{align}\label{eq:roughness_pre_defined}
    r_i = -\exp( \ln r_{t} - (\ln r_{t}-\ln r_{b})\frac{i-1}{k - 1}),
\end{align}
where $i\in [1,\cdots,k]$ denotes the index of basis. In testing, we empirically set the number of bases $k=12$,  $r_{t}=300$, and  $r_{b}=10$.

\paragraph{Pre-training light model}
The light model $L_\Psi(\cdot)$ is pre-trained on a public available synthetic dataset, Blobby and Sculpture datasets~\cite{chen2018ps}. We trained the $L_\Psi(\cdot)$ for 100 epoches, with batch size to be $64$. 
We adopt the Adam optimizer~\cite{kingma2014adam} for updating the network parameter $\Psi$ with learning rate $5.0 \times 10^{-4}$. 
The light network is pre-trained for once, based on the pre-train loss $\mathcal{L}_{\text{pre}} = (1- \vl^T \overline{\vl}) + (e - \overline{e})^2$, where the first term is cosine loss for light directions, the second term is mean-square-error for intensities.
The same network is then used for all other testing datasets at test time.

\paragraph{Testing}
At testing stage, we continue refining the lighting from pre-trained light net $L_\Psi(\cdot)$, while the proposed normal-MLP $N_\Theta(\cdot)$ and material-MLP $M_\Phi(\cdot)$ are optimized from scratch  via the reconstruction loss.
As the reconstruction loss, we use the mean absolute difference, which is the absolute difference between observed intensity $\mM \in \R^{p\times n} $ and re-rendered intensity $\overline{\mM}$.
\begin{align}
    \mathcal{L} = \frac{1}{pn} \sum_{i=1}^{p} \sum_{j=1}^{n} |\mM_{i,j} - \overline{\mM}_{i,j}| ,
\end{align}
where the above summation is over all $p$ pixels under $n$ different light sources.
At each iteration, we sampled pixels from $8$ images and feed them to the networks. The iterations per-epoch depends on the number of images of the scene. We run $2000$ epoches in total. 
The Adam optimizer is used with the learning rate being $10^{-3}$ for all parameters.

\paragraph{Training and testing time}
Our framework is implemented in PyTorch and runs on a single NVIDIA RTX3090 GPU.
The pre-training time of our light model $L_\Psi$ only takes around $2$ hours. In comparison, previous deep methods~\cite{chen2020learned,kaya2021uncalibrated} take more than $22$ hours in training light models. The reason is that we shift part of the burden of solving lightings from the neural light model to the inverse rendering procedural. Hence, our light model can be relatively lightweight and easy to train compared to previous deep learning based light estimation networks~\cite{chen2020learned,kaya2021uncalibrated}.

In testing, our method takes an average of $16$ minutes to process each of the ten objects in DiLiGenT~\cite{shi2016benchmark} benchmark, ranging from $13$ minutes to $21$ minutes.
In comparison, previous CNN-based inverse rendering methods~\cite{taniai2018neural,kaya2021uncalibrated} take on average $53$ minutes per object in testing. The reason is that both of our object modeling net $N_\Theta, M_\Phi$ are simple MLPs. Hence, we can achieve a much faster forward-backward time when optimizing the MLP-based network than previous CNN-based methods.

\begin{table}[t]
\parbox{.48\linewidth}{
\centering
\caption{Ablation study on effectiveness of progressive specular bases (PSB). We compare the models with and without progressive specular bases. Applying progressive specular bases will consistently improve estimation accuracy. 
}
\scriptsize
\begin{tabular}{@{}l|ccc@{}}
\toprule
Model                    & direction  & intensity    & normal \\ \midrule
$\vr$   & 5.30 & 0.0400 & 9.39   \\
$\vr$ $+$ PSB      & 4.42 &  0.0382 & 7.71   \\
trainable $\vr$  & 4.75 & 0.0372 & 8.57   \\ 
trainable $\vr$  $+$ PSB   & \textbf{4.02} & \textbf{0.0365} & \textbf{7.05}   \\\bottomrule
\end{tabular}
\label{tab:specular_bases_r}
}
\hfill
\parbox{.48\linewidth}{
\centering
\caption{Quantitative results on DiLiGenT where our model takes different lighting as initialization. Our model performs consistently well when varying levels of noise are applied to the lighting estimation.
It shows that our model is robust against the errors in lighting estimation.}
\scriptsize
\begin{tabular}{@{}c|ccccc@{}}
\toprule
      & $L_\Psi$ & $+20^{\circ}$ & $+30^{\circ}$ & $+50^{\circ}$ & $+70^{\circ}$ \\ \midrule
direction    & 4.02   & 4.07     & 4.14     & 4.34     & 4.40     \\
intensity    & 0.0365 & 0.0337   & 0.0356   & 0.0358   & 0.0355   \\
normal & 7.05   & 7.40     & 7.40     & 7.40     & 7.44     \\ \bottomrule
\end{tabular}
\label{tab:init_error}
}
\end{table}

\section{Experiments}
\subsection{Testing Dataset}
We conduct experiments in following public real-world datasets: DiLiGenT~\cite{shi2016benchmark},  Gourd\&Apple dataset~\cite{alldrin2008photometric}, and Light Stage Data Gallery~\cite{chabert2006relighting}.
They all provide calibrated light directions and light intensities as ground truth for evaluation.
DiLiGenT contains $10$ objects; each object has $96$ images captured under different lighting conditions; a high-end laser scanner captured ground truth surface normal is available for evaluation.
Gourd\&Apple contains three objects; each object has around $100$ images captured under different lighting conditions;
Light Stage Data Gallery contains $252$ images per object; following previous works~\cite{chen2019self}, we select in total $133$ images illuminated by forward-facing lights for photometric stereo. 
Unfortunately, Gourd\&Apple and Light Stage Data Gallery did not provide ground truth surface normal for quantitative evaluation.

\subsection{Evaluation metrics}
In this paper, we use mean angular errors (MAE) as an evaluation metric for surface normal estimate and light direction estimation. Lower MAE is preferred.

As light intensities among different images can only be estimated up to a scale factor, we follow previous work~\cite{chen2019self} to use scale-invariant relative error
\begin{align}
    E_{si} = \frac{1}{n} \sum_i^n \frac{|se_i - \overline{e_i}|}{\overline{e_i}},
\end{align}
where $e_i, \overline{e_i}$ denote the estimated and ground truth light intensity of $i$-th light respectively;  $s$ is the scale factor computed by solving $\text{argmin}_s \sum_i^n(se_i-\overline{e_i})^2$ with least squares. Lower scale-invariant relative error is preferred.

\begin{SCfigure}[][t]
\centering
\caption{Visualization of the lighting optimization under noised input. 
The predicted light distribution over a sphere is represented as the spheres above.
The left-most lighting sphere shows the noised lighting estimation at the start of optimization. 
Our model gradually refines the incorrect lighting during optimization (from left to right) and provides the optimized result.
}
\includegraphics[width=0.5\textwidth]{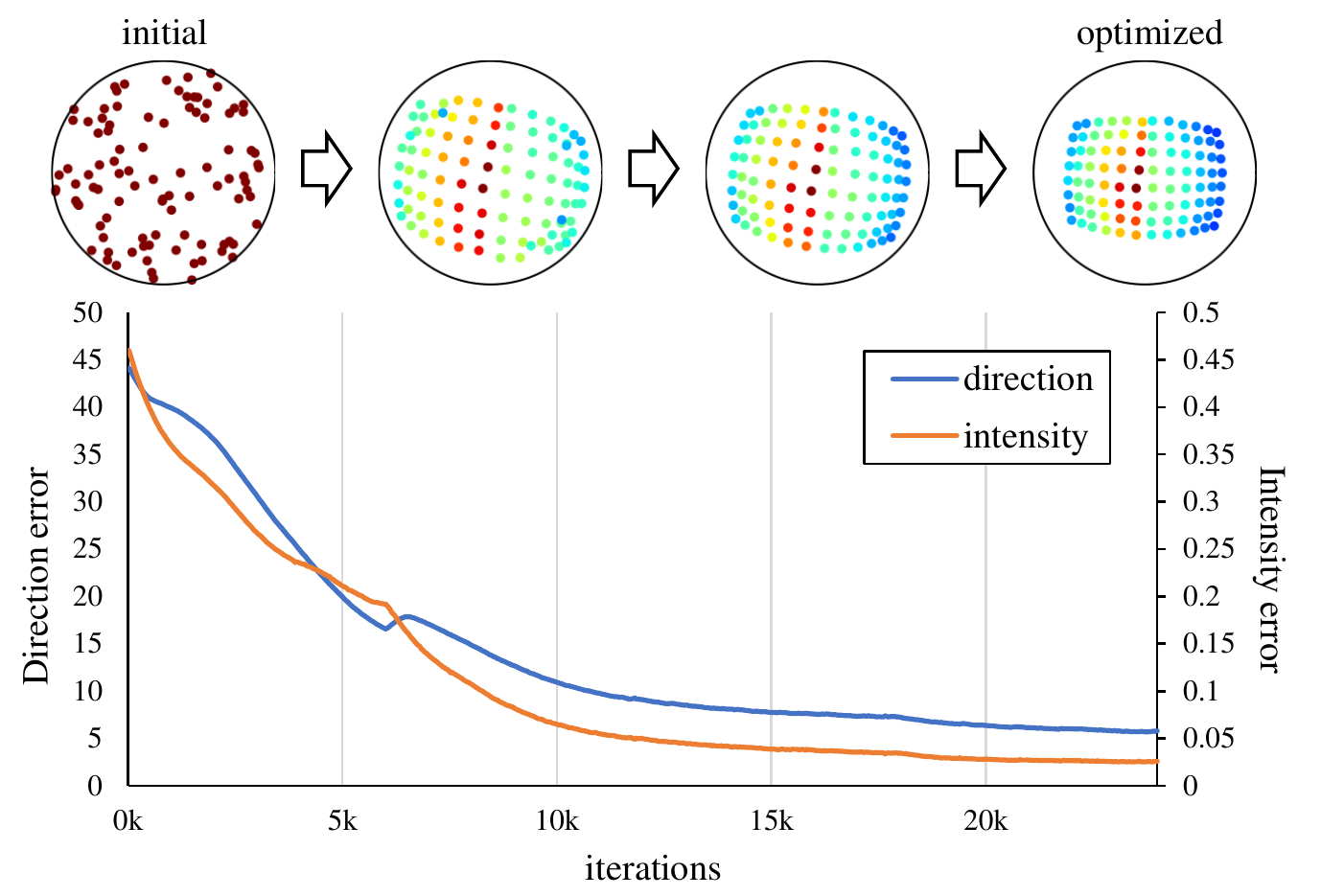}
\label{fig:init_error}
\end{SCfigure}

\subsection{Ablation study}
\paragraph{Effectiveness of the Progressive Specular Bases}
To show the effectiveness of the proposed progressive specular bases, we conduct the ablation study as shown in \cref{tab:specular_bases_r}. The models are evaluated in the DiLiGenT dataset.
In the first and second row of the table, we evaluate the model with and without the progressive specular bases. It shows that the model with progressive specular bases achieves lower error in both lighting and normal estimation.
In the third and fourth row of the table, instead of using fixed (as defined by \eqref{eq:roughness_pre_defined}) roughness terms, we set these roughness terms as trainable parameters. Results also demonstrate that when the roughness terms are trainable, progressive specular bases consistently improve the estimation accuracy. We also observe that the trainable roughness with progressive specular bases achieves the best performance. Our analysis is that, by relaxing the training of specular roughness, the network can adjust these terms for more accurate material estimation. Hence, it also leads to more accurate lighting and normals.

\begin{table}[t]
\centering
\caption{Evaluation results on DiLiGenT benchmark. Here, \textbf{bold} indicates the best results and {\ul underline} denotes the second best results. }
\begin{subtable}{\textwidth}
\centering
\caption{Normal estimation results on DiLiGenT benchmark. }
{\scriptsize
\begin{tabular}{@{}l|cccccccccc|c@{}}
\toprule
Method                                                                                & Ball          & Bear          & Buddha        & Cat           & Cow           & Goblet        & Harvest        & Pot1          & Pot2          & Reading        & average       \\ \midrule
SM10\cite{shi2010self}                                               & 8.90          & 11.98         & 15.54         & 19.84         & 22.73         & 48.79         & 73.86          & 16.68         & 50.68         & 26.93          & 29.59         \\
WT13\cite{wu2013calibrating}                                         & 4.39          & 6.42          & 13.19         & 36.55         & 19.75         & 20.57         & 55.51          & 9.39          & 14.52         & 58.96          & 23.93         \\
PF14\cite{papadhimitri2014closed}                                    & 4.77          & 9.07          & 14.92         & 9.54          & 19.53         & 29.93         & 29.21          & 9.51          & 15.90         & 24.18          & 16.66         \\
LC18\cite{lu2017symps}                                               & 9.30          & 10.90         & 19.00         & 12.60         & 15.00         & 18.30         & 28.00          & 12.40         & 15.70         & 22.30          & 16.30         \\
UPS-FCN\cite{chen2018ps}                                             & 6.62          & 11.23         & 15.87         & 14.68         & 11.91         & 20.72         & 27.79          & 13.98         & 14.19         & 23.26          & 16.02         \\
BK21\cite{kaya2021uncalibrated}                                      & 3.78          & 5.96          & 13.14         & 7.91          & 10.85         & 11.94         & 25.49          & 8.75          & 10.17         & 18.22          & 11.62         \\
SDPS-Net\cite{chen2019self}                                          & 2.77          & 6.89          & {\ul 8.97}    & 8.06          & 8.48          & 11.91         & 17.43          & 8.14          & 7.50          & 14.90          & 9.51          \\
SK21\cite{sarno2021neural}                                           & 3.46          & {\ul 5.48}    & 10.00         & 8.94          & {\ul 6.04}    & 9.78          & 17.97          & 7.76          & {\ul 7.10}    & 15.02          & 9.15          \\
GCNet\cite{chen2020learned}+PS-FCN\cite{chen2018ps} & {\ul 2.50}    & 5.60          & \textbf{8.60} & {\ul 7.90}    & 7.80          & {\ul 9.60}    & {\ul 16.20}    & {\ul 7.20}    & {\ul 7.10}    & {\ul 14.90}    & {\ul 8.70}    \\
Ours                                                                                  & \textbf{1.24} & \textbf{3.82} & 9.28          & \textbf{4.72} & \textbf{5.53} & \textbf{7.12} & \textbf{14.96} & \textbf{6.73} & \textbf{6.50} & \textbf{10.54} & \textbf{7.05} \\ \bottomrule
\end{tabular}
}
\label{tab:diligent_normal}
\end{subtable}
\begin{subtable}{\textwidth}
\centering
\caption{Light intensity estimation results on DiLiGenT benchmark. }
{\scriptsize
\begin{tabular}{@{}l|cccccccccc|c@{}}
\toprule
Method & Ball            & Bear            & Buddha          & Cat             & Cow             & Goblet          & Harvest         & Pot1            & Pot2            & Reading         & average         \\ \midrule
PF14\cite{papadhimitri2014closed}   & 0.0360          & 0.0980          & 0.0530          & {\ul 0.0590}    & 0.0740          & 0.2230          & 0.1560          & \textbf{0.0170} & \textbf{0.0440} & 0.1220          & 0.0882          \\
LCNet\cite{chen2019self}  & 0.0390          & {\ul 0.0610}    & 0.0480          & 0.0950          & 0.0730          & 0.0670          & 0.0820          & 0.0580          & {\ul 0.0480}    & 0.1050          & 0.0676          \\
GCNet\cite{chen2020learned}  & {\ul 0.0270}    & 0.1010          & {\ul 0.0320}    & 0.0750          & \textbf{0.0310} & {\ul 0.0420}    & {\ul 0.0650}    & 0.0390          & 0.0590          & {\ul 0.0480}    & {\ul 0.0519}    \\
Ours   & \textbf{0.0194} & \textbf{0.0186} & \textbf{0.0206} & \textbf{0.0321} & {\ul 0.0621}    & \textbf{0.0418} & \textbf{0.0230} & {\ul 0.0303}    & 0.0816          & \textbf{0.0352} & \textbf{0.0365} \\ \bottomrule
\end{tabular}
}
\label{tab:diligent_light_intensity}
\end{subtable}
\begin{subtable}{\textwidth}
\centering
\caption{Light direction estimation results on DiLiGenT benchmark. }
{\scriptsize
\begin{tabular}{@{}l|cccccccccc|c@{}}
\toprule
Method & Ball          & Bear          & Buddha        & Cat           & Cow           & Goblet        & Harvest       & Pot1          & Pot2          & Reading       & average       \\ \midrule
PF14\cite{papadhimitri2014closed}   & 4.90          & 5.24          & 9.76          & 5.31          & 16.34         & 33.22         & 24.99         & 2.43          & 13.52         & 21.77         & 13.75         \\
LCNet\cite{chen2019self}  & 3.27          & 3.47          & 4.34          & \textbf{4.08} & {\ul 4.52}    & 10.36         & {\ul 6.32}    & 5.44          & {\ul 2.87}    & \textbf{4.50} & 4.92          \\
GCNet\cite{chen2020learned}  & {\ul 1.75}    & {\ul 2.44}    & \textbf{2.86} & 4.58          & \textbf{3.15} & {\ul 2.98}    & \textbf{5.74} & \textbf{1.41} & \textbf{2.81} & {\ul 5.47}    & \textbf{3.32} \\
Ours   & \textbf{1.43} & \textbf{1.56} & {\ul 4.22}    & {\ul 4.41}    & 4.94          & \textbf{2.26} & 6.41          & {\ul 3.46}    & 4.19          & 7.34          & {\ul 4.02}    \\ \bottomrule
\end{tabular}
}
\label{tab:diligent_light_direction}
\end{subtable}
\label{tab:diligent_benchmark}
\end{table}

\paragraph{Robustness on light modeling}
As stated above, our model shifts part of the burden of solving lightings from the neural light modeling to the later inverse rendering procedural. Hence, even if our pre-trained light model $L_\Psi$ does not provide perfect lighting estimations, our later procedural can continue refining its estimation via the reconstruction error. To demonstrate the robustness of our model against the errors of the light model, we conduct the experiments where different levels of noise are added to the lightings, as shown in \cref{tab:init_error} and \cref{fig:init_error}.  In \cref{tab:init_error}, the first column shows the results of our model on DiLiGenT with the pre-trained $L_\Psi$. From the second column, different levels of noise (noise that is up to certain degrees) are applied to the lightings. The light directions are randomly shifted, and the light intensities are all re-set to ones. Then, we further refined this noised lighting estimation via the inverse rendering procedural at the testing stage. As illustrated in \cref{tab:init_error}, even when the light directions are randomly shifted up to $70$ degrees, our model still achieves comparable performance after the optimization. In \cref{fig:init_error}, we visualize how our model gradually refines the lighting estimation during the course of optimization.

\subsection{Evaluation on DiLiGenT benchmark}

\begin{SCfigure}[][t]
\centering
\includegraphics[width=0.55\textwidth]{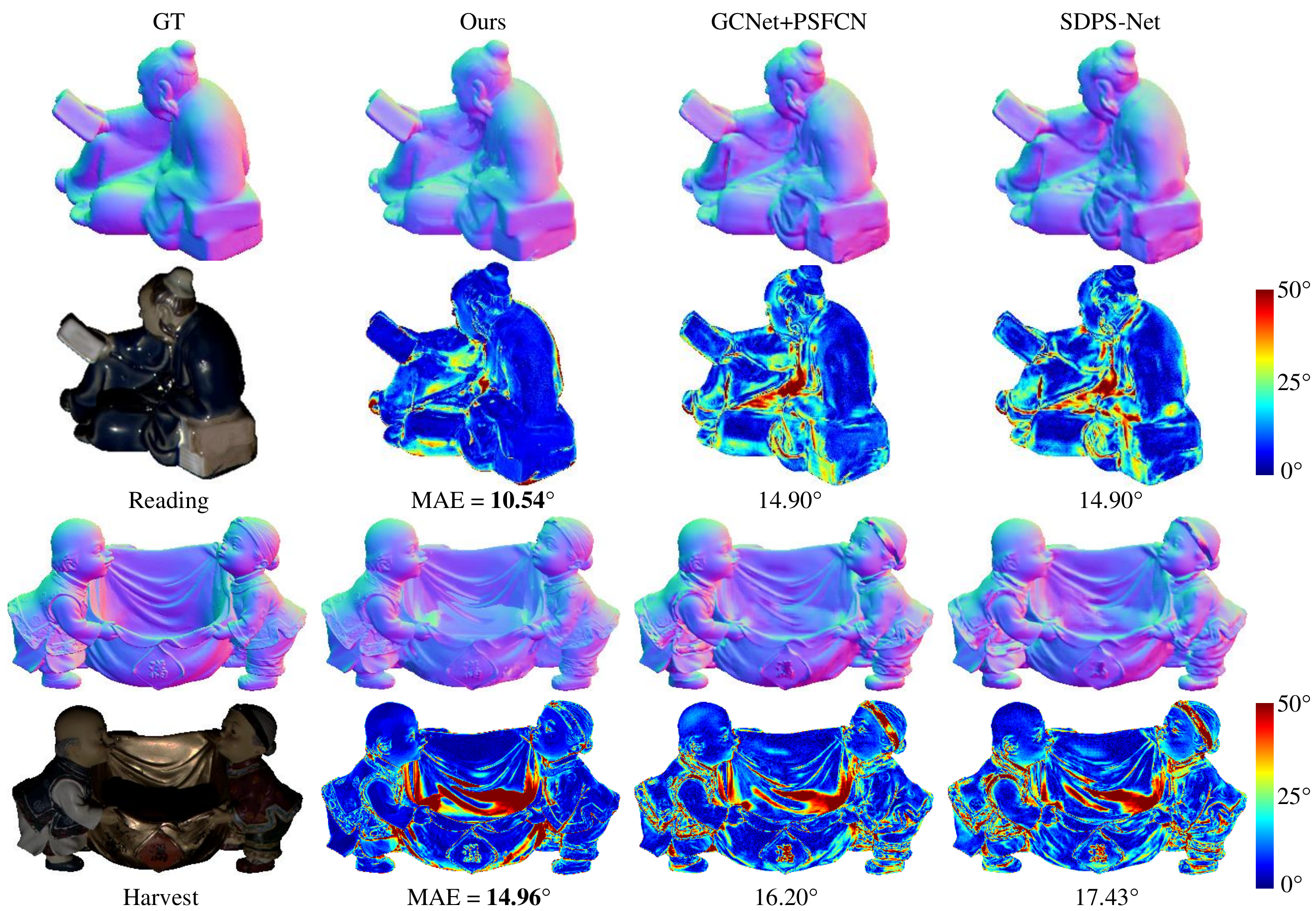}
\caption{Visualized comparisons of normal estimation for ``Reading'' and ``Harvest'' in DiLiGenT. Our method produces better normal estimation than others, particularly in regions with specularities (\eg see the head of ``reading'', the golden sack of ``Harvest'').}
\label{fig:diligent_normal}
\end{SCfigure}

\paragraph{Results on normal estimation} 
We evaluate our method on the challenging DiLiGenT benchmark and compare our method with previous works.
The quantitative result on normal estimation is shown in \cref{tab:diligent_normal}.  
We achieve the best average performance, and we outperform the second-best method by $1.65$ degrees on average.
Thanks to the proposed progressive specular bases, our method performs particularly well on those objects with specularities. There are a large number of specularities in ``Reading'' and ``Goblet'', where our method outperforms the others by $4.36$ and $2.48$ degrees, respectively. \Cref{fig:diligent_normal} shows qualitative comparison on ``Reading'' and ``Harvest'. Our method produces much better results on those specular regions.

\paragraph{Results on light estimation}
The quantitative results on light intensity estimation are shown in \cref{tab:diligent_light_intensity}. Our model achieves the best performance on average, which is $0.0365$ in relative error. The results on light direction estimation are presented in \cref{tab:diligent_light_direction}, where we also demonstrate a comparable result to previous methods.  \Cref{fig:light_results} showcases the visualization of the lighting results. As LCNet~\cite{chen2019self} discretely represents the light direction into bins, their estimation looks very noisy (see lighting in ``Reading''). In contrast, our model can continuously refine the lights. Hence, our lighting estimation preserves smoother pattern overall.

\subsection{Evaluation on other real world dataset}
\begin{figure}[t]
\centering
\includegraphics[width=0.99\textwidth]{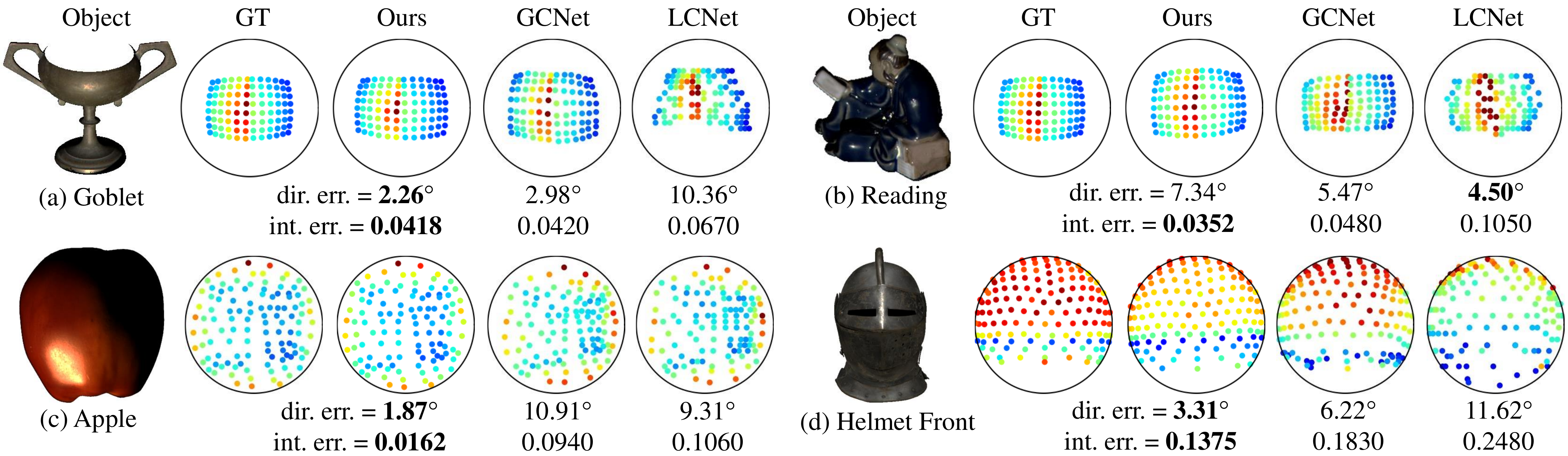}
\caption{Visualized comparisons of the ground-truth and estimated lighting distribution for the DiLiGenT dataset, Gourd\&Apple dataset, and Light Stage dataset. It demonstrates that our lighting estimation is also robust in different datasets under different lighting distributions.
}
\label{fig:light_results}
\end{figure}

We then evaluate our method on other challenging real-world datasets. \cref{tab:apple} shows that our method achieves the best performance in lighting estimation in the Gourd\&Apple dataset. In \cref{fig:light_results}, We visualized the estimated lighting in ``Apple'' from Gourd\&Apple dataset, and ``Helmet Front'' from Light Stage dataset. These results manifest that our method can also reliably recover the lighting in different light distributions. Our specular modeling is also applicable to different datasets under different materials. Please refer to supplementary material for more results.

\begin{table}[t]
\centering
\caption{Evaluation results on Gourd\&Apple dataset.}
\begin{subtable}{0.5\textwidth}
\centering
\caption{Results on light intensity.}
\scriptsize
\begin{tabular}{@{}l|ccc|c@{}}
\toprule
Method & Apple           & Gourd1          & Gourd2          & Avg.            \\ \midrule
PF14\cite{papadhimitri2014closed}  & {0.1090} & {0.0960}  & {0.3290} & 0.1780  \\
LCNet\cite{chen2019self}  & 0.1060          & 0.0480          & \textbf{0.1860} & 0.1130          \\
GCNet\cite{chen2020learned}  & 0.0940          & 0.0420          & 0.1990          & 0.1120          \\
Ours   & \textbf{0.0162} & \textbf{0.0272} & 0.2330          & \textbf{0.0921} \\ \bottomrule
\end{tabular}
\end{subtable}
\begin{subtable}{0.45\textwidth}
\centering
\caption{Results on light direction.}
\scriptsize
\begin{tabular}{@{}l|ccc|c@{}}
\toprule
Method & Apple         & Gourd1        & Gourd2        & Avg.          \\ \midrule
PF14\cite{papadhimitri2014closed} & {6.68}   & {21.23}   & {25.87}    &17.92  \\
LCNet\cite{chen2019self}  & 9.31          & 4.07          & 7.11          & 6.83          \\
GCNet\cite{chen2020learned}  & 10.91         & 4.29          & 7.13          & 7.44          \\
Ours   & \textbf{1.87} & \textbf{2.34} & \textbf{2.01} & \textbf{2.07} \\ \bottomrule
\end{tabular}
\end{subtable}
\label{tab:apple}
\end{table}

\section{Discussions and Conclusions}\label{sec:discussions}
In this paper, we propose a neural representation for lighting and surface normal estimation via inverse rendering. The surface is explicitly modeled with diffuse and specular components, and the GBR ambiguity is resolved by fitting on these photometric cues. To avoid local minima during optimization, we propose \emph{progressive specular bases} for fitting the specularities.
Our method provides state-of-the-art performance on lighting estimation and shape recovery on challenging real-world datasets.

\textbf{Limitations and future work:} 
The inter-reflections, subsurface scattering, and image noises are not considered in our image rendering \eqref{eq:our_image_rendering}. Our model may fail if these terms are prominent on an object's surface. Explicitly modeling these terms and jointly refining them within the same framework will be an intriguing direction to pursue.

\textbf{Acknowledgments}
This research is funded in part by ARC-Discovery grants (DP190102261 and DP220100800), a gift from Baidu RAL, as well as a Ford Alliance grant to Hongdong Li.

%
%
\bibliographystyle{splncs04}
\bibliography{egbib}

\appendix
\section*{Supplementary Material}
\section{Implementation Details}
\begin{figure}
\centering
\begin{subfigure}[b]{0.45\textwidth}
\centering
\includegraphics[width=0.9\textwidth]{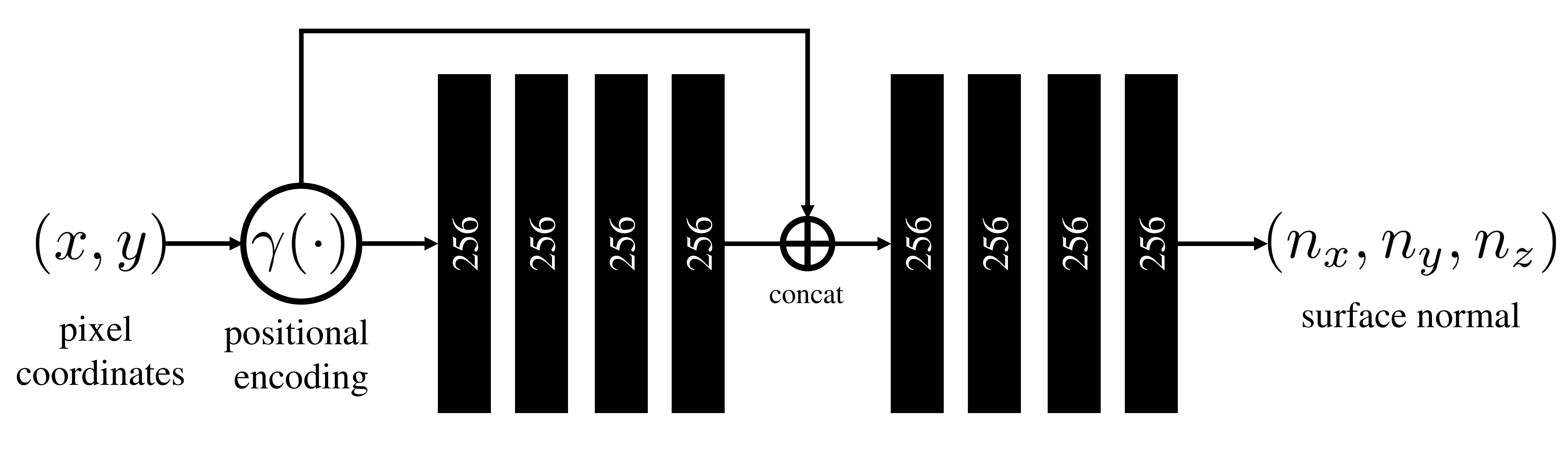}
\caption{Normal network.}
\label{fig:normalnet}
\end{subfigure}
\begin{subfigure}[b]{0.45\textwidth}
\centering
\includegraphics[width=0.9\textwidth]{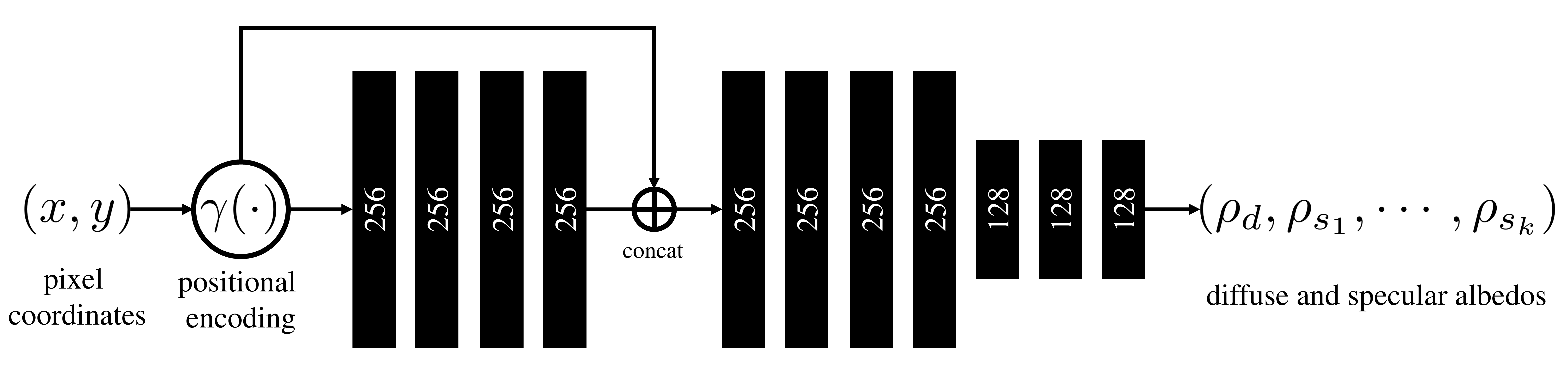}
\caption{Material network.}
\label{fig:materialnet}
\end{subfigure}
\begin{subfigure}[b]{0.8\textwidth}
\centering
\includegraphics[width=0.99\textwidth]{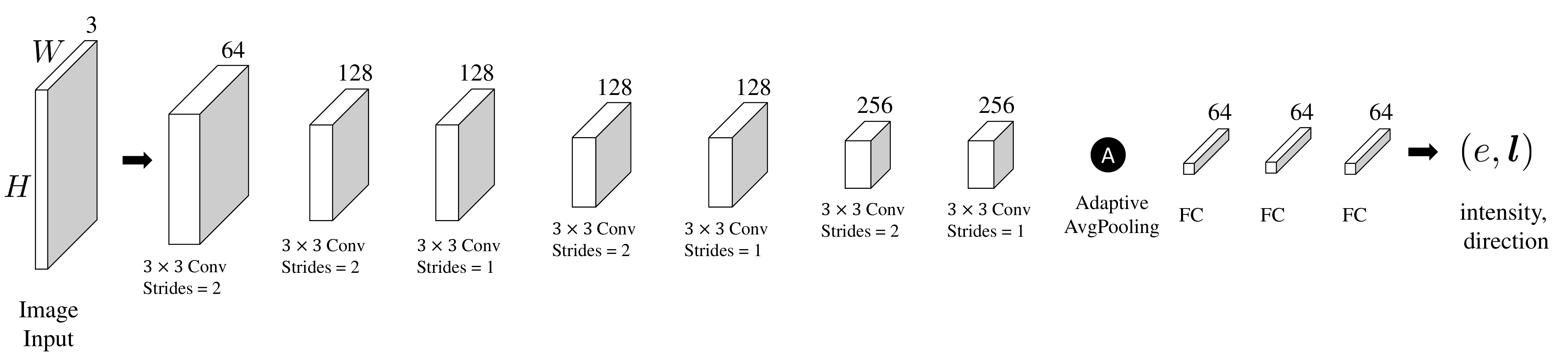}
\caption{Lighting network.}
\label{fig:lightnet}
\end{subfigure}
\caption{The network architectures of our networks.}
\label{fig:network}
\end{figure}

\paragraph{Network architectures}
Here, we describe our network architectures in detail.
In \cref{fig:normalnet}, the surface normal net $N_\Theta(\cdot)$ uses $8$ fully-connected layers with $256$ channels. It takes positional encoded~\cite{mildenhall2020nerf} pixel coordinates $\gamma(x),\gamma(y)$ as input, directly output the surface normal at that position $\vn =[n_x,n_y,n_z]^T$.
As shown in \cref{fig:materialnet}, the material net $M_\Phi(\cdot)$ uses the same structure but with $3$ more  fully-connected layers than normal net. It takes the same  positional encoded pixel coordinates input and outputs the diffuse and specular albedos of the surface point.
As shown in \cref{fig:lightnet}, the lighting network $L_\Psi (\cdot)$ consists of $7$ convolutional ReLU layers and $3$ fully connected layers. It takes the image with size $H\times W\times 3$ as  input, directly outputs the light intensity $e$ and light direction $\vl$ of that image.
\paragraph{Early supervision} Following previous works~\cite{barron2014shape,li2022neural}, we additionally use the surface smoothness constraints and shape-from-contour priors as the early supervision in our network. After early-stage training in the first half iterations, we discard these priors and train the network via photometric loss.

\newpage

\section{Visualization of the effectiveness of PSB}
\begin{figure}[!h]
\centering
\includegraphics[width=0.99\textwidth]{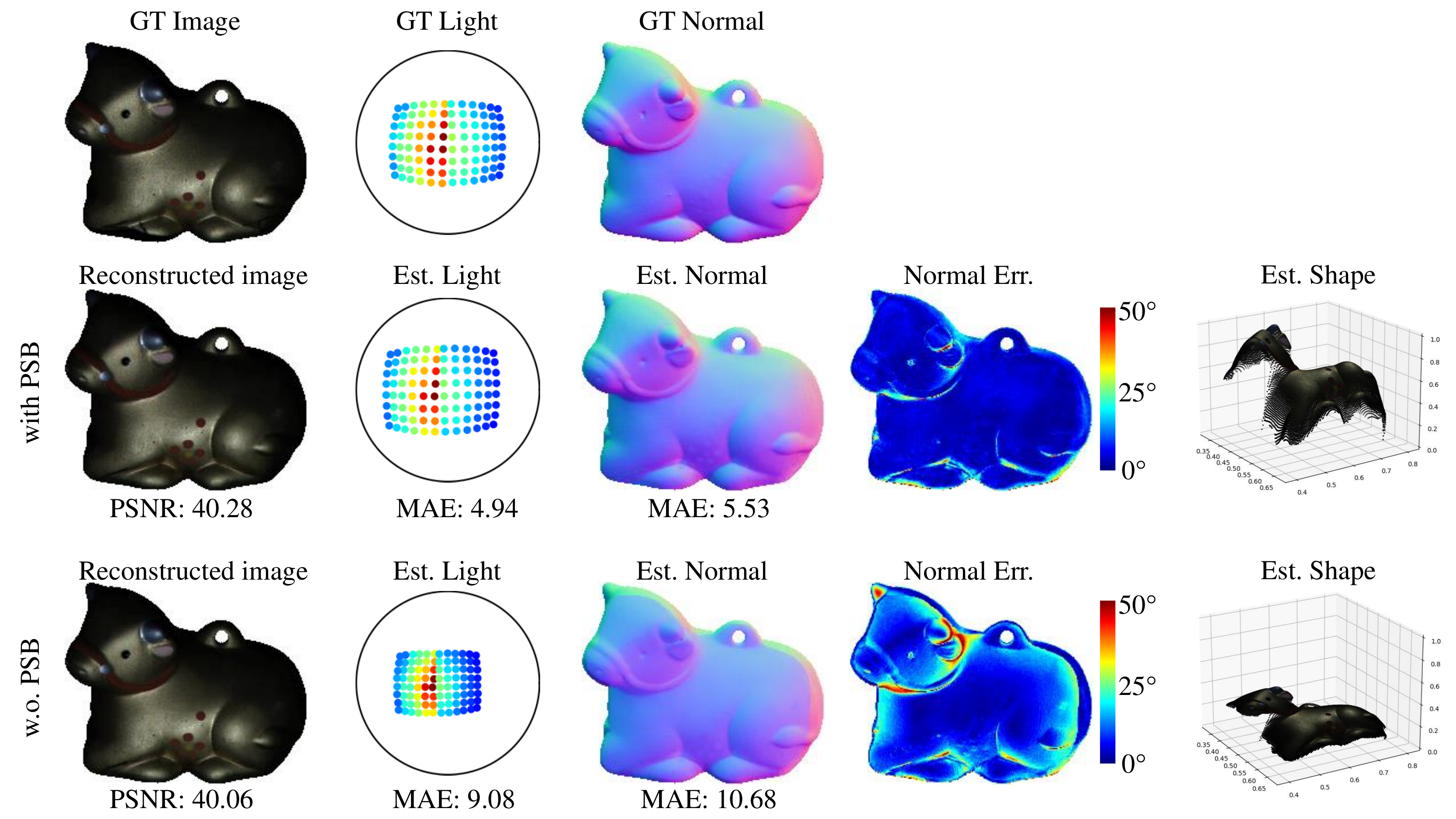}
\caption{Visualized comparisons of with/without using the \emph{progressive specular basis} (PSB).}

\label{fig:visual_PSB}
\end{figure}

Recall that, although the GBR ambiguity can be reduced up to a binary concave/convex ambiguity under our model, there is no guarantee that no local minima exist during the optimization. To effectively avoid the local minimas during the optimization, we propose the progressive specular bases (PSB) for the network.
In \cref{fig:visual_PSB}, we provide the visual comparison between the model with PSB and the model without PSB. The first row displays the observed ground truth image under a light source, the ground truth light distribution, and the ground truth surface normal. The second row and third row display the reconstructed image, the estimated light distribution, the estimated normal,  the error map of estimated normal, and the estimated shape from our ``with PSB model'' and ``without PSB model'' respectively. 

As we can see, the ``without PSB'' produces a worse light and normal estimation. Both the light and normal are ``shifted'' along the $z$ axis. However, its reconstructed image still presents a similar quality to the observed ground truth (PSNR: $40.06$dB). This observation coincides with the observation from Belhumeur~\cite{belhumeur1999bas}, where they also observed that the differences in shape are hard to be discerned from the frontal images given a small scale along the $z$ axis. 

The PSB can provide prior information to the network and limit the space of possible solutions by forcing the network to fit on the shiny specularities first in the early stage of optimization. Hence, by applying with the PSB, even with a poor network initialization, our network can still effectively avoid the local minimas to achieve better results.

\newpage
\section{Ablation study on lighting model}
\begin{figure}[!h]
\centering
\includegraphics[width=0.99\textwidth]{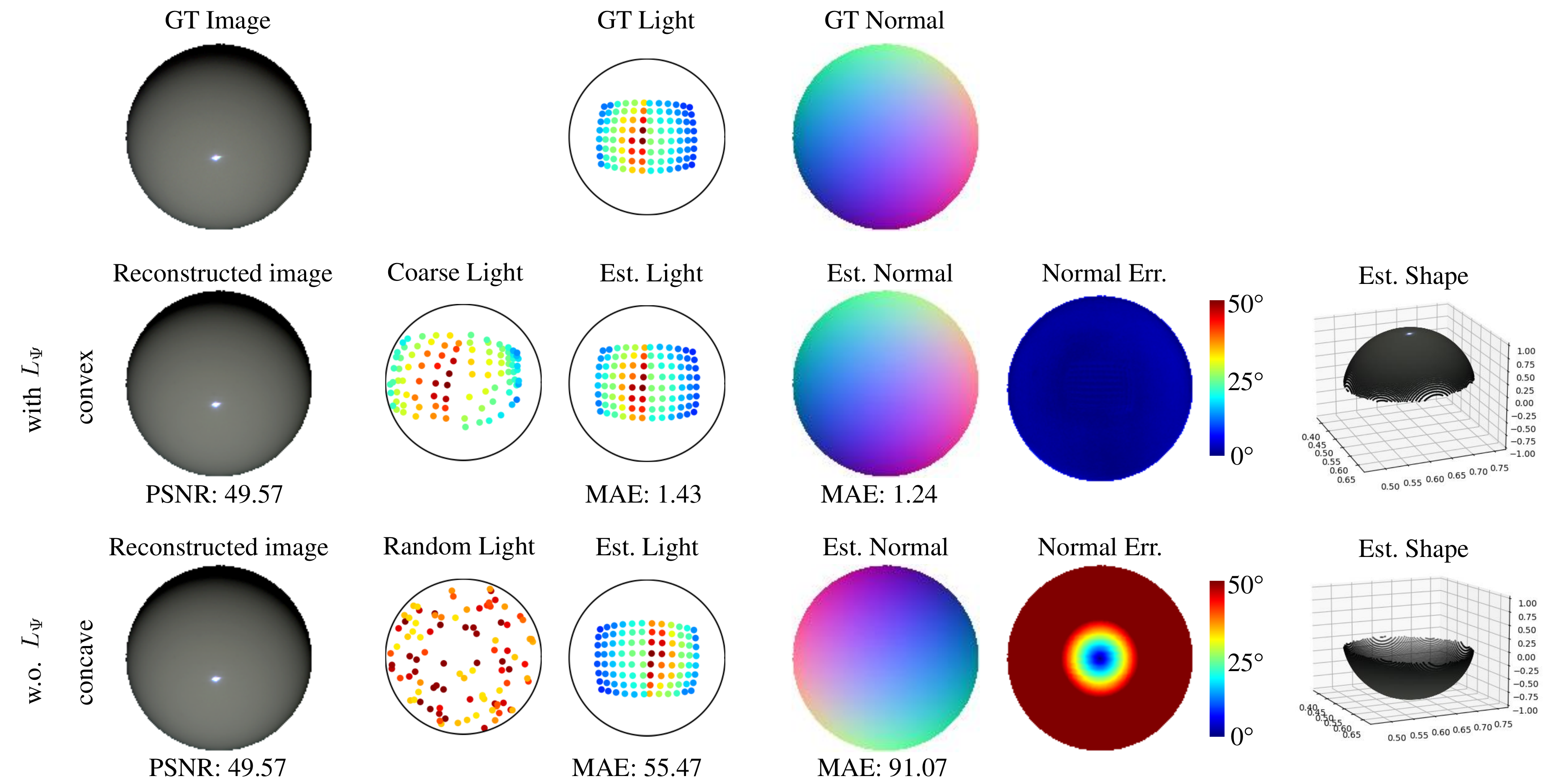}
\caption{Visualization of the effectiveness of lighting model. }
\label{fig:convex_ambiguity}
\end{figure}

In this section, we conduct two experiments to showcase the effectiveness of the lighting network. As shown in \cref{fig:convex_ambiguity}: on the first row, we showcase the observed(GT) image, ground truth lights and normals; on the second row, we display reconstructed image and estimated result using the model with lighting network $L_\Psi(\cdot)$; on the third row, we present results using the model without lighting network and takes randomized lights as initialization.

Without using the lighting network, we take randomized lights as initialization. Our network may sometimes produce a flipped surface as a result, as shown in the third row in \cref{fig:convex_ambiguity}. As we can see in the second row and third row in \cref{fig:convex_ambiguity}, the estimated lights and normals are flipped in the $x,y$ axis. In the third row, the mean angular error (MAE) for light direction is $55.47$ degrees, and normal error is $91.07$ degrees. However, its reconstructed image is almost identical to the observed ground truth image.

During the experiments, we observed that this convex/concave ambiguity can be easily resolved by providing the model with a coarse lighting estimation, as shown in second row in \cref{fig:convex_ambiguity}.
Our lighting model $L_\Psi(\cdot)$ can provide a coarse lighting estimation as the starting point, which is sufficient to for the followed self-supervised network to further refine the coarse results and produce the correct lights.

\newpage
\section{Robustness on Sparse Inputs}
In this section, we present the results on DiLiGenT~\cite{shi2016benchmark} dataset with only 16 images at the inputs.
Following previous works on sparse inputs for photometric stereo~\cite{li2019learning}, we selected 16 images as input for our method and others for comparison. The errors are shows in \cref{tab:sparse_input}.
As we can see from the table, our method still outperform the state-of-the-art with only 16 images. Besides, with 16 images as input, our method only drop $0.72$ degrees in MAE in normal estimation, while GCNet\cite{chen2020learned}+PSFCN\cite{chen2018ps} drops $2.04$ degrees in MAE. It also demonstrate that our method is robust against sparse input.

\begin{table}[h]
\centering
\caption{Quantitative comparison on DiLiGenT with only 16 images at input.}
{
\begin{subtable}{0.7\textwidth}
\centering
\caption{MAE of surface normal.}
\begin{tabular}{@{}c|cc@{}}
\toprule
model & All images & 16 images \\ \midrule
Ours & 7.05 & 7.77 \\
GCNet\cite{chen2020learned}+PSFCN\cite{chen2018ps} & 8.70 & 10.74 \\ \bottomrule
\end{tabular}
\label{tab:sparse_input_nor}
\end{subtable}
\begin{subtable}{0.7\textwidth}
\centering
\caption{Scale-invariant relative error of light intensities.}
\begin{tabular}{@{}c|cc@{}}
\toprule
model & All images & 16 images \\ \midrule
Ours & 0.0365 & 0.0548 \\
GCNet\cite{chen2020learned} & 0.0519 & 0.0550 \\ \bottomrule
\end{tabular}
\label{tab:sparse_input_li}
\end{subtable}
\begin{subtable}{0.7\textwidth}
\centering
\caption{MAE of light directions.}
\begin{tabular}{@{}c|cc@{}}
\toprule
model & All images & 16 images \\ \midrule
Ours & 4.02 & 5.02 \\
GCNet\cite{chen2020learned} & 3.32 & 4.04 \\ \bottomrule
\end{tabular}
\label{tab:sparse_input_ld}
\end{subtable}
}
\label{tab:sparse_input}
\end{table}

\section{Results on DiLiGenT benchmark}
\begin{figure}[!h]
\centering
\includegraphics[height=0.93\textheight]{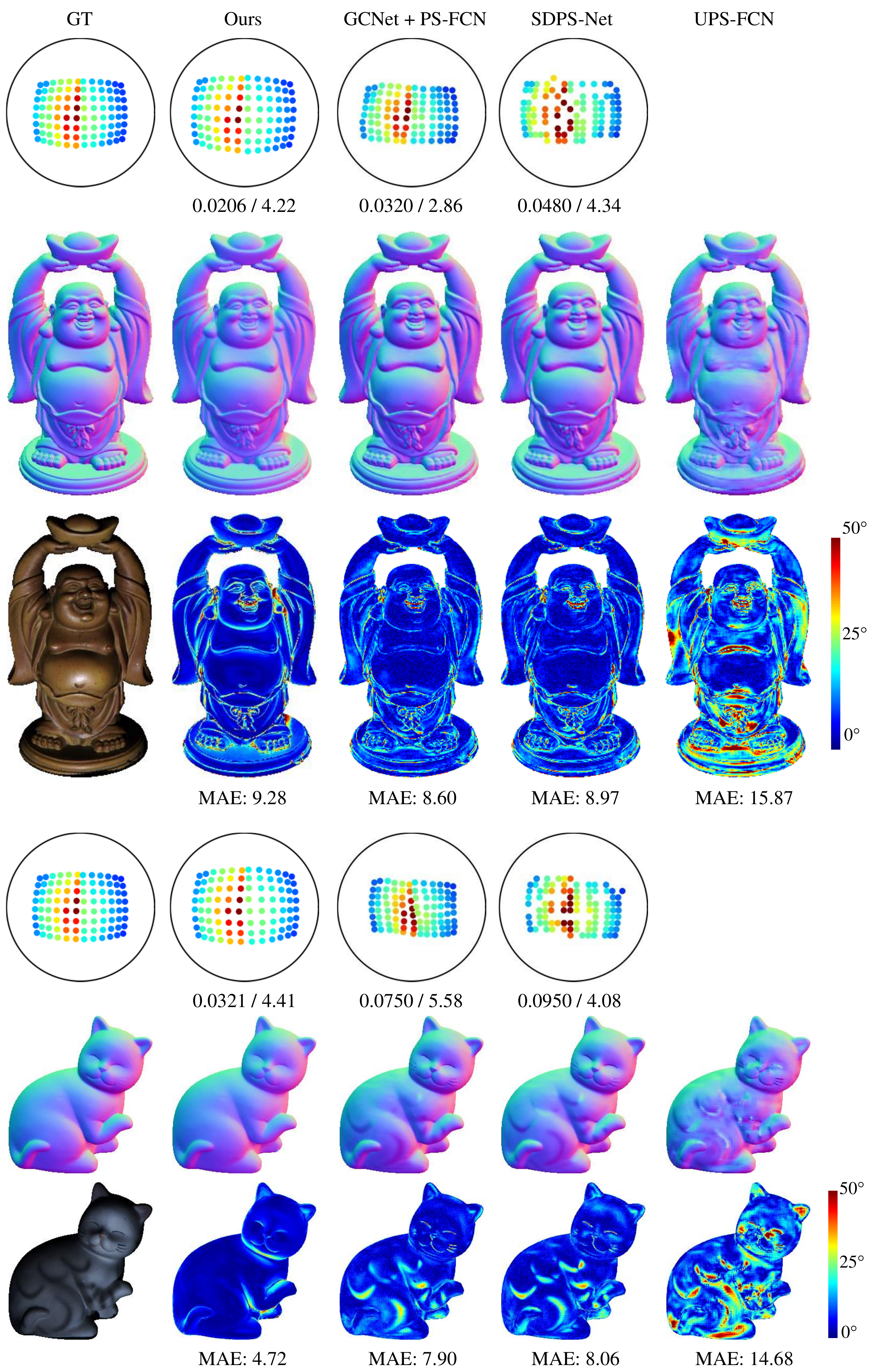}
\caption{Results for ``Buddha'' and ``Cat'' from DiLiGenT dataset. }
\label{fig:buddha_cat}
\end{figure}

\begin{figure}[!h]
\centering
\includegraphics[height=0.93\textheight]{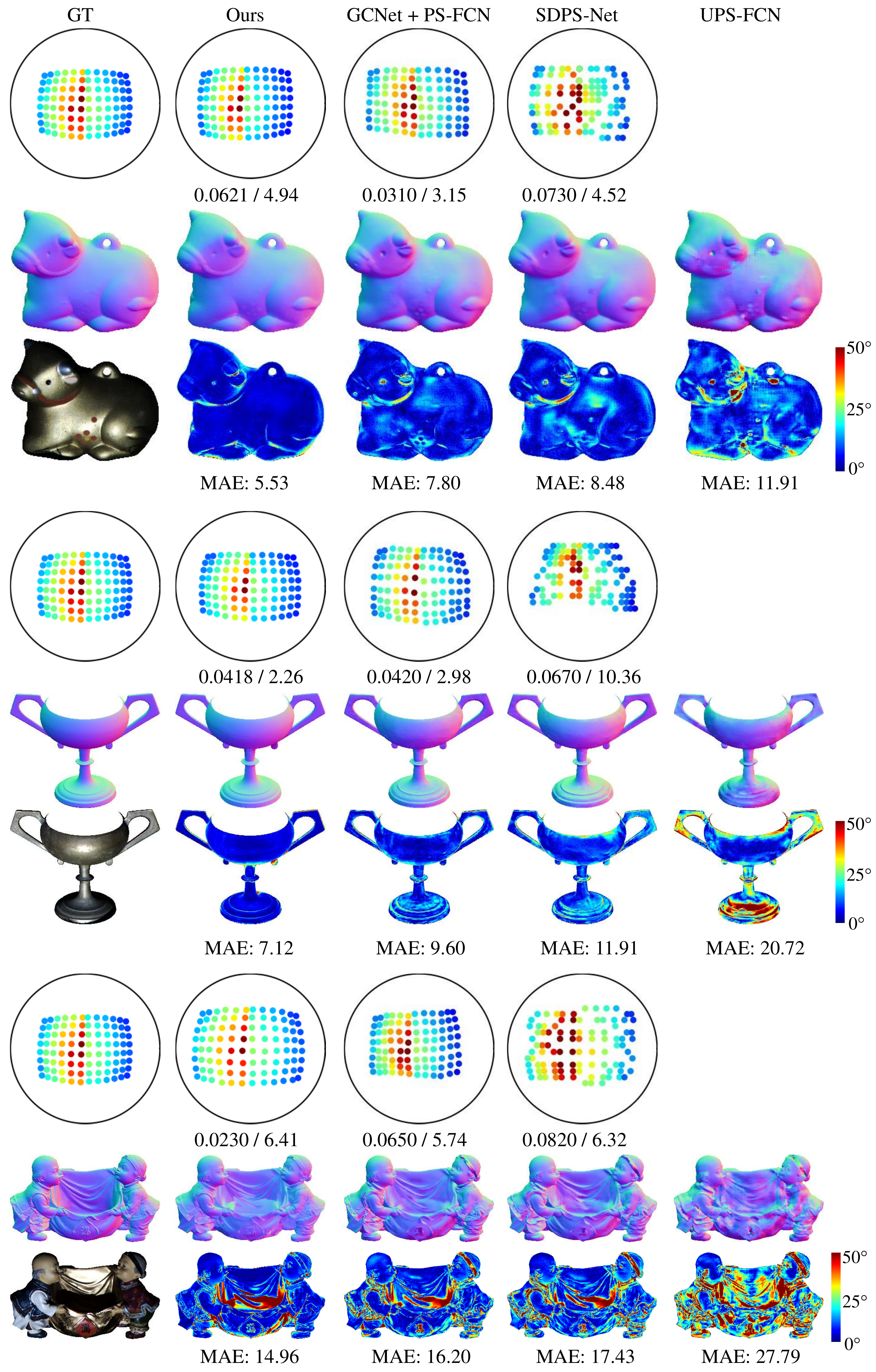}
\caption{Results for ``Cow'' , ``Goblet'', and ``Harvest'' from DiLiGenT dataset. }
\label{fig:cow_go_har}
\end{figure}

\begin{figure}[!h]
\centering
\includegraphics[height=0.93\textheight]{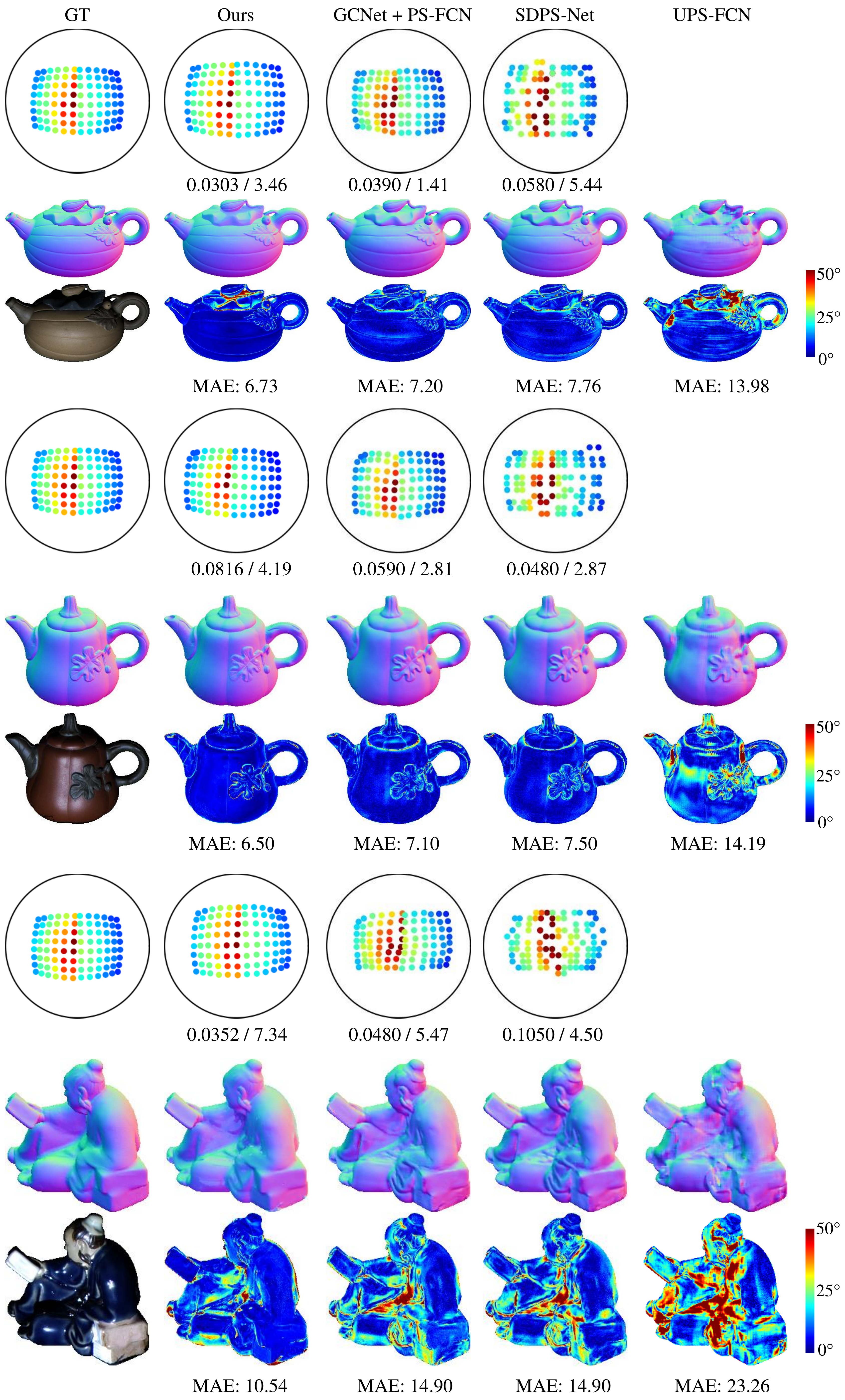}
\caption{Results for ``Pot1'' , ``Pot2'', and ``Reading'' from DiLiGenT dataset. }
\label{fig:pot_reading}
\end{figure}

In this section, we present the results on DiLiGenT~\cite{shi2016benchmark} dataset, as shown in the following \cref{fig:buddha_cat}, \ref{fig:cow_go_har} and \ref{fig:pot_reading}.
For each object, the first row displays the ground truth lighting, ours estimated lighting, and lighting results from GCNet~\cite{chen2020learned} and SDPS-Net~\cite{chen2019self}. The second row displays the ground truth surface normal and estimated surface normal by ours and competing methods.
The last row display the observed image and the error map of the estimated surface normal. We also present the quantitative evaluation for lighting and normal below the lighting and error map. Note that UPS-FCN~\cite{chen2018ps} can not estimate the lighting.  

\begin{figure}
\centering
\includegraphics[width=0.8\textwidth]{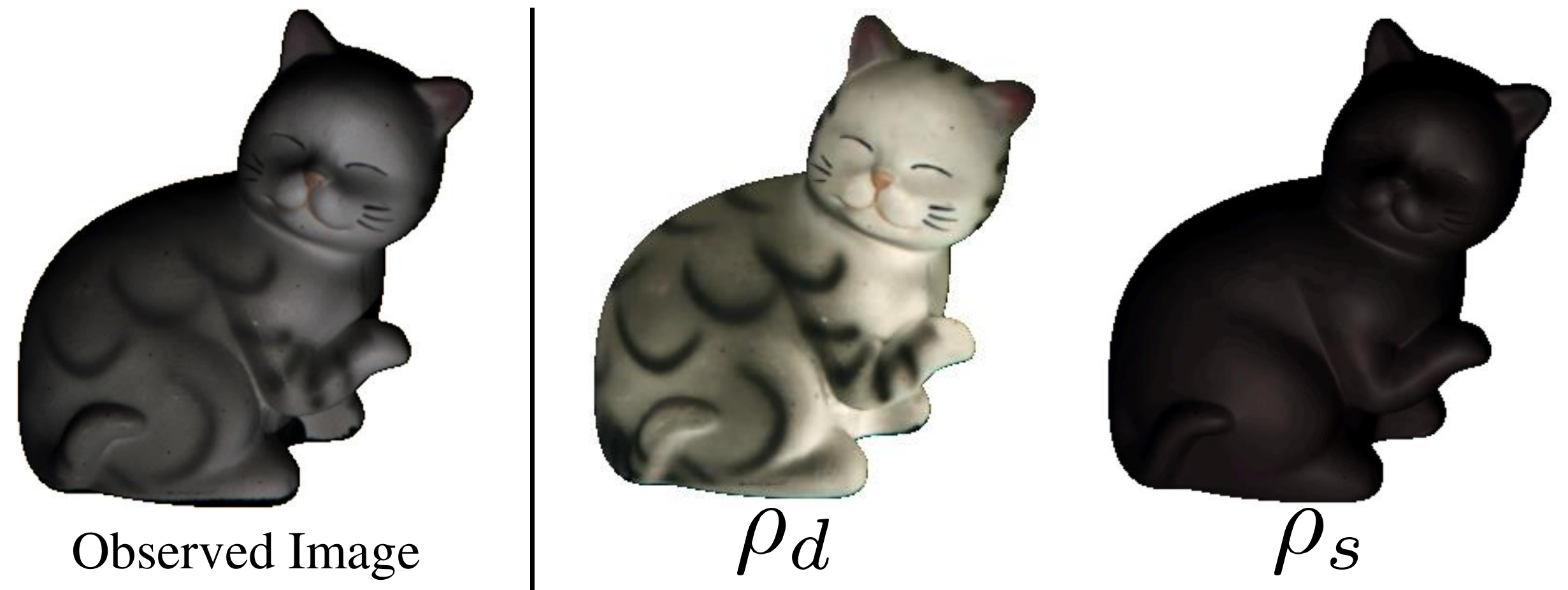}
\caption{Visualization of reconstructed $\rho_d$ and $\rho_s$ in object ``Cat''.
}
\label{fig:cat_diff_spec}
\end{figure}

\paragraph{Results on almost Lambertian surface} As we can see from the results, our method works well for specular objects, as well as objects that appear to be very diffuse, such as ``Cat''.
In order to better understand why our method also works well on objects like ``Cat'', we visualized the reconstructed terms $\rho_d$ and $\rho_s$ in \cref{fig:cat_diff_spec}. \Cref{fig:cat_diff_spec} shows that the ``Cat'' is not purely diffuse and contains very soft specularities. Our method is able to capture and use these soft specularities as clues for estimating the surface normal. 

\section{Results on Apple\&Gourd dataset}
\begin{figure}[!h]
\centering
\includegraphics[height=0.80\textheight]{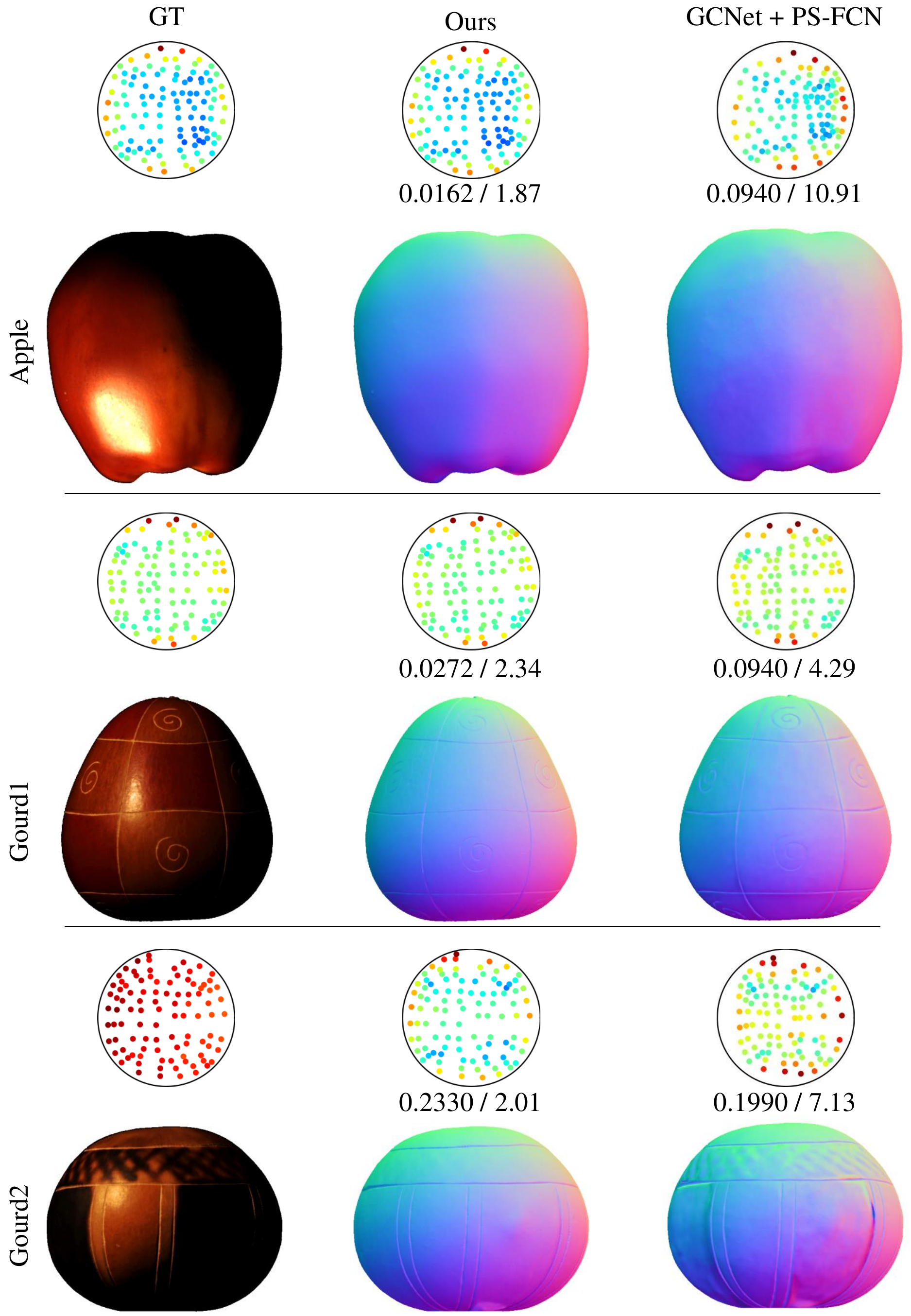}
\caption{Results for ``Apple'' , ``Gourd1'', and ``Gourd2'' from Apple\&Gourd dataset. }
\label{fig:apple}
\end{figure}

In this section, we present the results on Apple\&Gourd~\cite{alldrin2008photometric} dataset. In \cref{fig:apple}, for each object, the first row displays the ground truth lighting, ours estimated lighting, and lighting results from GCNet~\cite{chen2020learned}. The second row displays the observed image and estimated surface normal by ours and competing methods. Note that there is no ground truth surface normal available in this dataset, so we only visualized compare the normal results. As shown in ``Gourd2'', it is clear that our estimated normal present higher quality than previous state-of-the-art method GCNet~\cite{chen2020learned}+PSFCN~\cite{chen2018ps}.

\section{Results on synthetic dataset with 100 MERL BRDFs}
To evaluation our method across different surface materials and BRDFs, we test our method on a publicly available synthetic dataset\footnote{\url{https://github.com/guanyingc/UPS-GCNet}}: GCNet-Synthetic~\cite{chen2020learned}. The dataset consists of two rendered synthetic objects: Dragon and Armadillo for testing. This dataset was rendered with 100 MERL~\cite{Matusik:2003} BRDFs under 82 random light directions using physically based renderer Mitsuba\footnote{\url{http://mitsuba-renderer.org/}}. 

\begin{figure}
\centering
\begin{subfigure}{0.99\textwidth}
\centering
\includegraphics[width=0.37\textwidth]{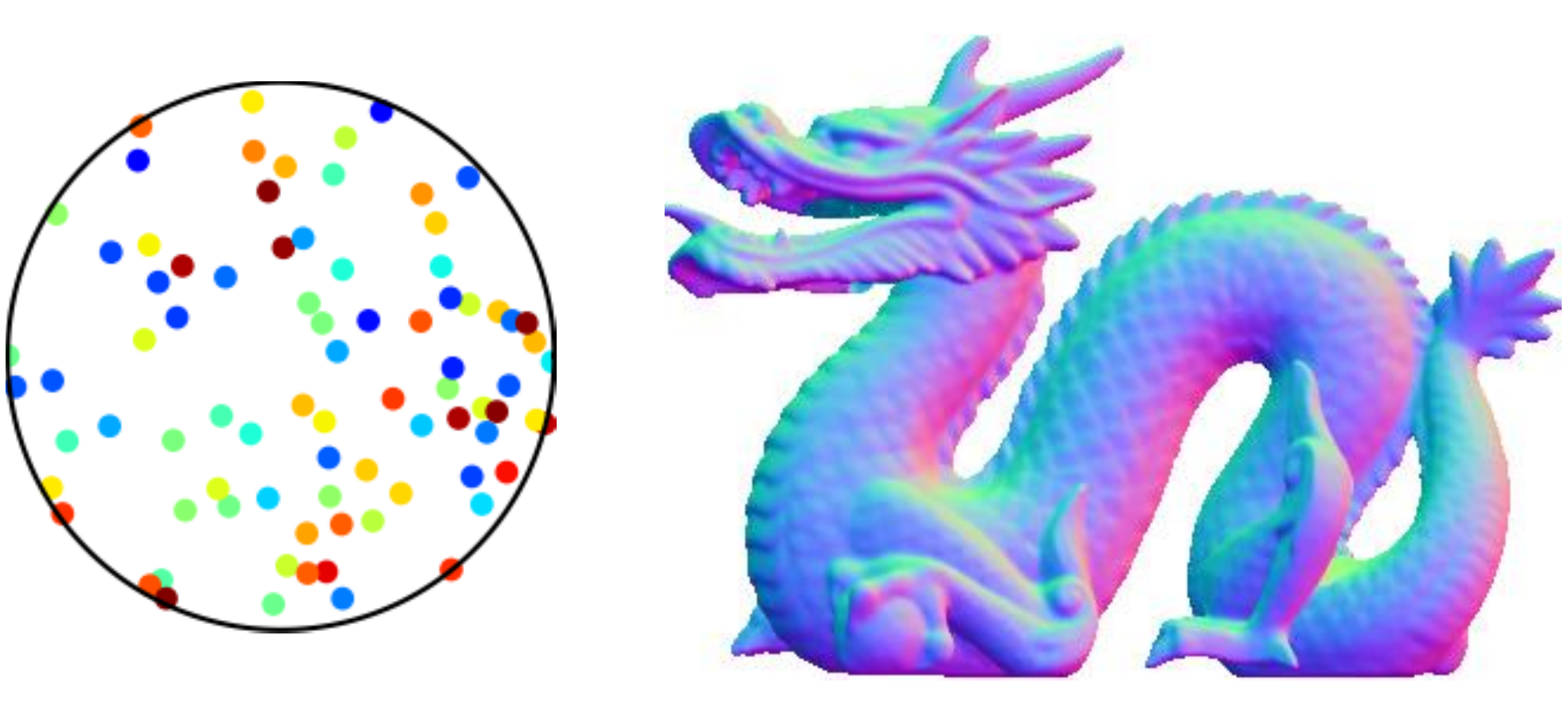}
\caption{Ground truth of lights and surface normals.
}
\label{fig:arma-ld}
\end{subfigure}

\begin{subfigure}{0.99\textwidth}
\centering
\includegraphics[width=0.99\textwidth]{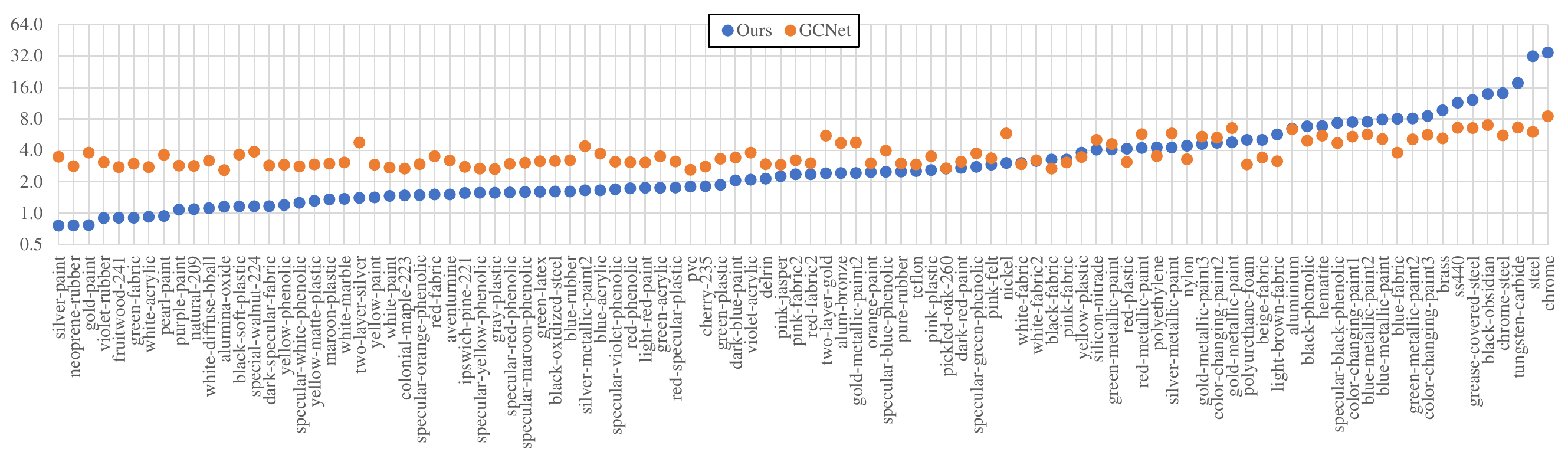}
\caption{MAE of light directions.  
}
\label{fig:dragon-ld}
\end{subfigure}

\begin{subfigure}{0.99\textwidth}
\centering
\includegraphics[width=0.99\textwidth]{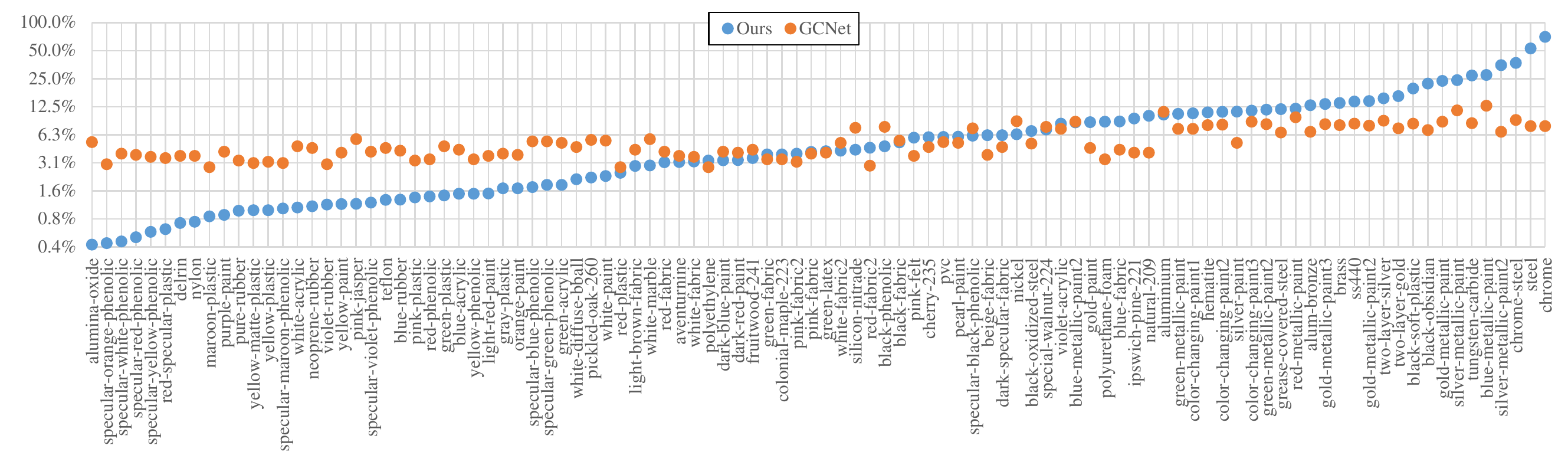}
\caption{Scale-invariant relative error in percentage of light intensities.
}
\label{fig:dragon-li}
\end{subfigure}

\begin{subfigure}{0.99\textwidth}
\centering
\includegraphics[width=0.99\textwidth]{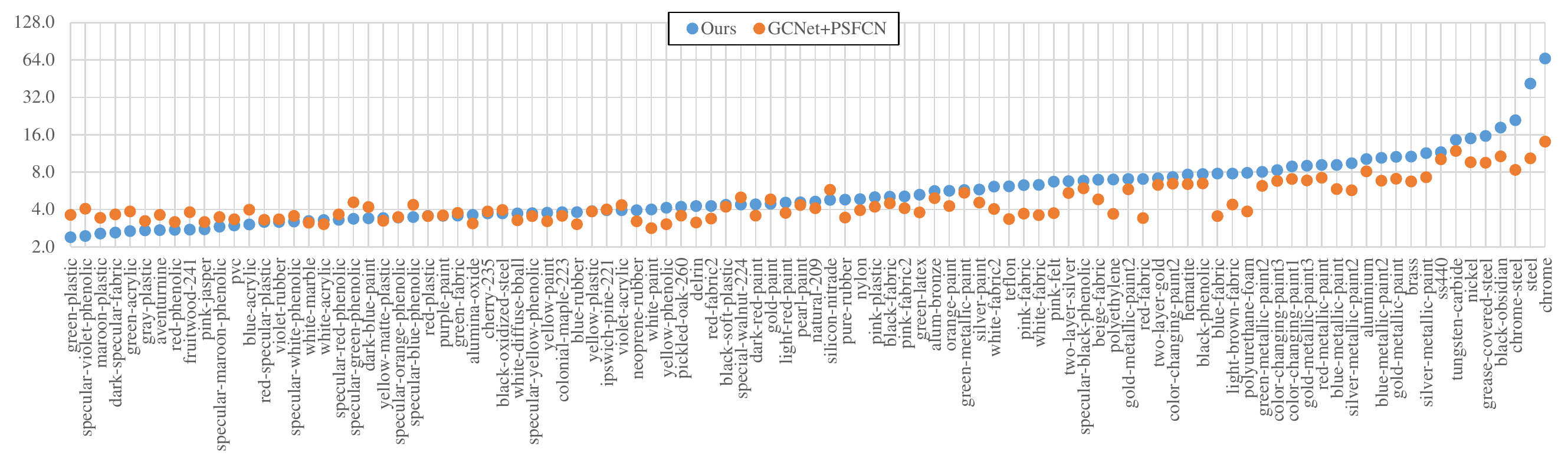}
\caption{MAE of surface normals.
}
\label{fig:dragon-nor}
\end{subfigure}
\caption{Comparison on object ``Dragon'' rendered with 100 MERL BRDFs.}
\label{fig:dragon}
\end{figure}

\begin{figure}
\centering
\begin{subfigure}{0.99\textwidth}
\centering
\includegraphics[width=0.3\textwidth]{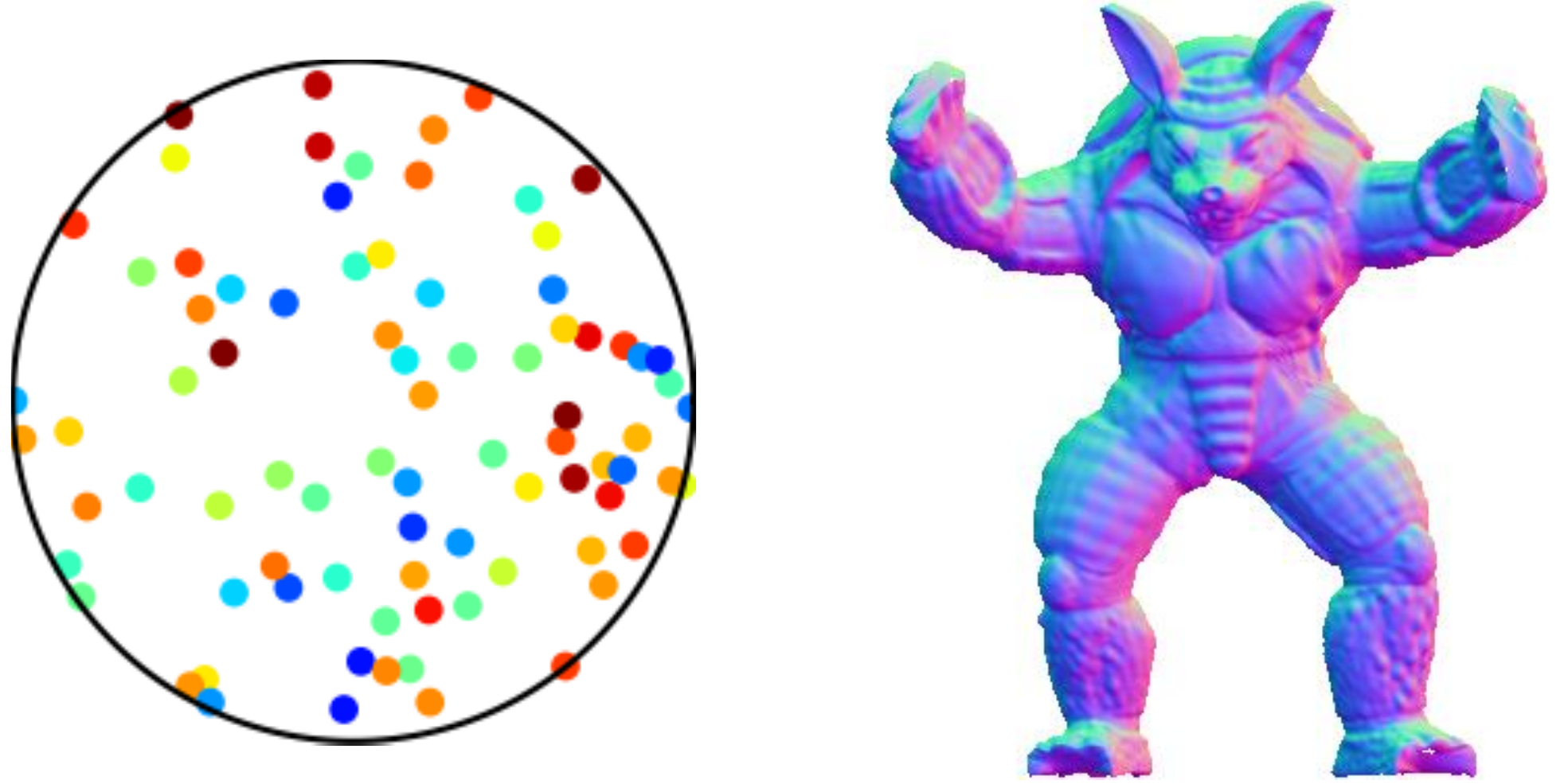}
\caption{Ground truth of lights and surface normals.
}
\label{fig:arma-ld}
\end{subfigure}

\begin{subfigure}{0.99\textwidth}
\centering
\includegraphics[width=0.99\textwidth]{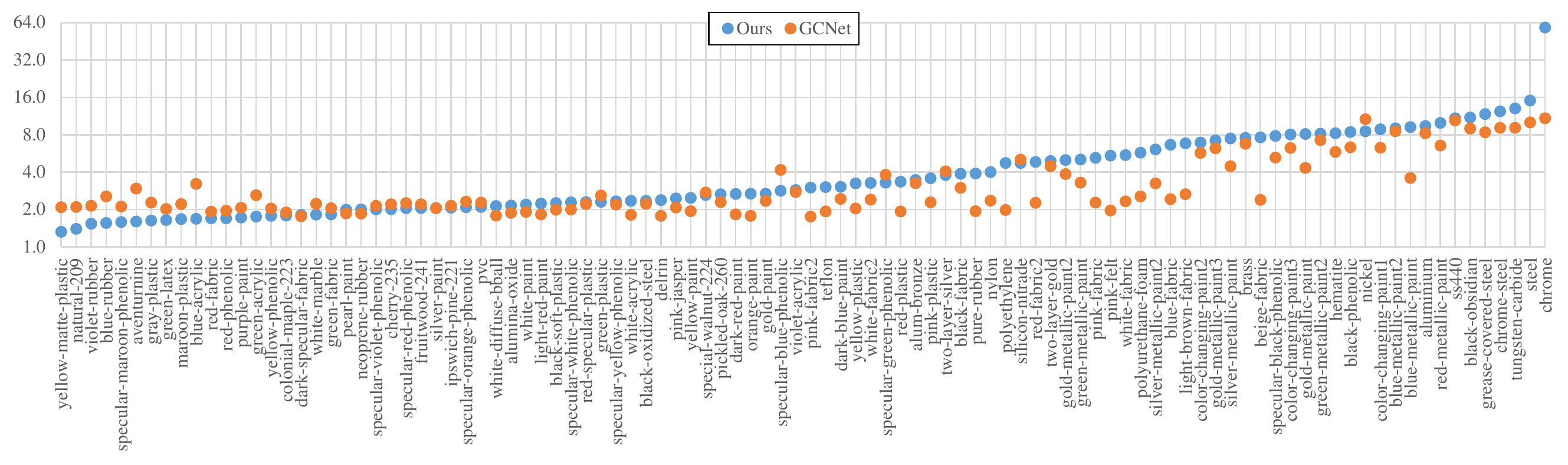}
\caption{MAE of light directions.  
}
\label{fig:arma-ld}
\end{subfigure}

\begin{subfigure}{0.99\textwidth}
\centering
\includegraphics[width=0.99\textwidth]{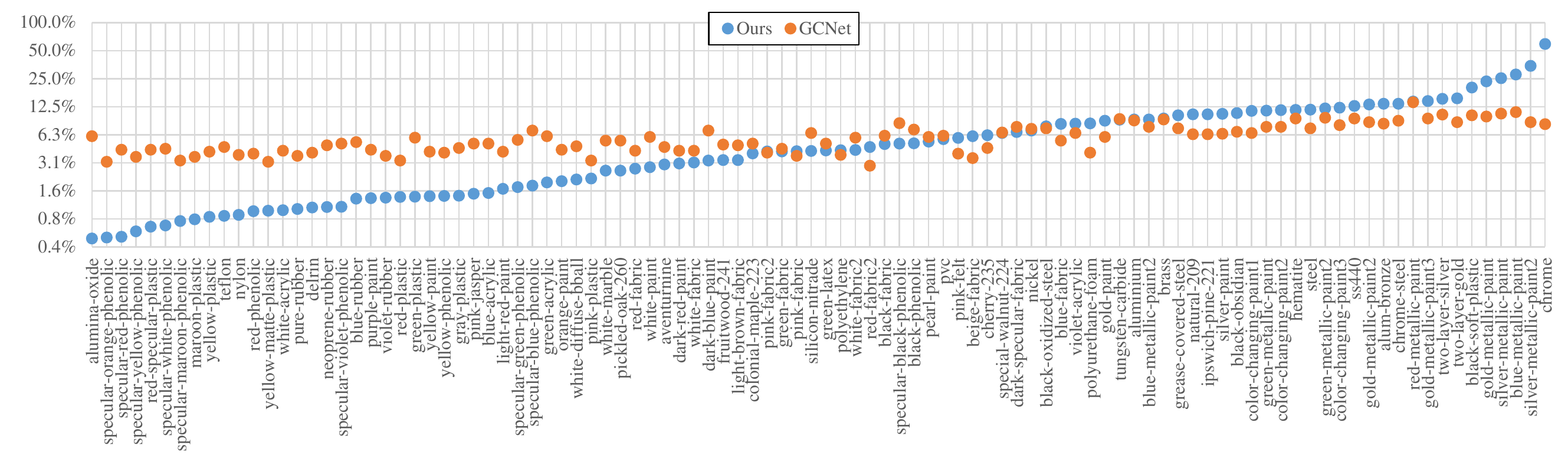}
\caption{Scale-invariant relative error in percentage of light intensities.
}
\label{fig:arma-li}
\end{subfigure}

\begin{subfigure}{0.99\textwidth}
\centering
\includegraphics[width=0.99\textwidth]{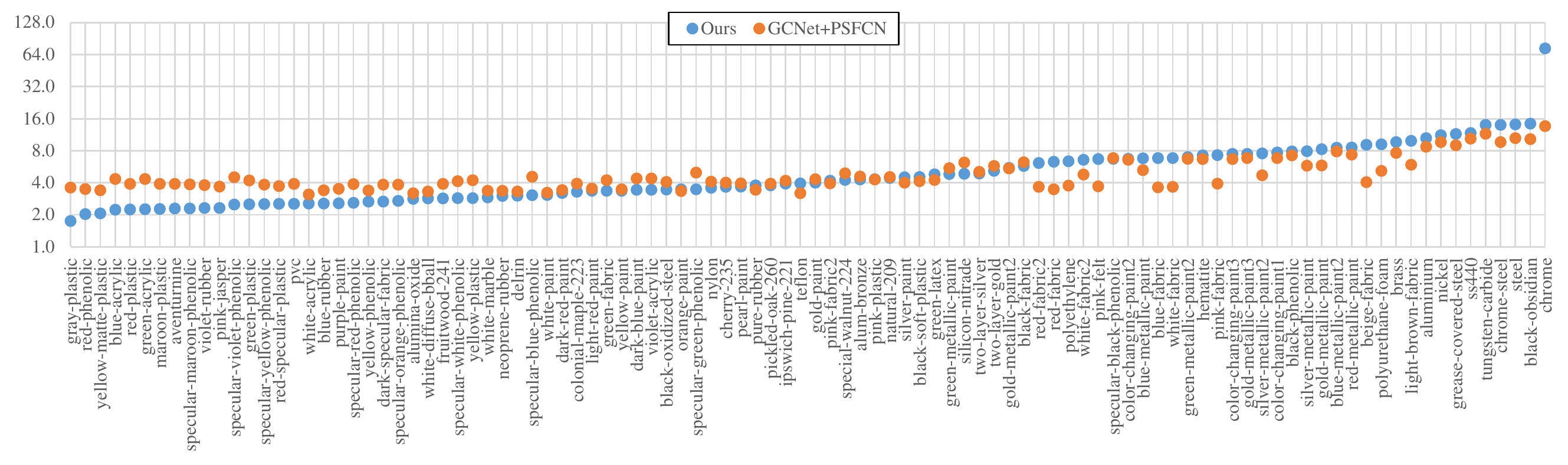}
\caption{MAE of surface normals.
}
\label{fig:arma-nor}
\end{subfigure}
\caption{Comparison on object ``Armadillo'' rendered with 100 MERL BRDFs.}
\label{fig:arma}
\end{figure}

We showcase the results in \cref{fig:dragon} and \cref{fig:arma}. As we can see from the figures, our method produce comparable results to GCNet~\cite{chen2020learned}. 

\begin{figure}[!h]
\centering
\includegraphics[width=0.35\textwidth]{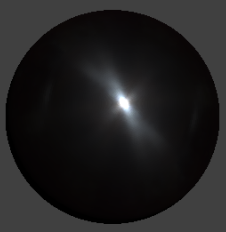} \includegraphics[width=0.36\textwidth]{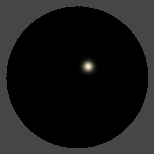}
\caption{Rendered sphere of ``steel''. Left is the data from MERL. Right is our estimated result.
}
\label{fig:merl}
\end{figure}

We dive into the MERL dataset and found that our method fails to fit the materials such as ``steel'', ``chrome'', and ``chrome-steel'', where they generally present asymmetric highlights as shown in \cref{fig:merl}.
Prior work~\cite{burley2012physically} believed that these anomaly asymmetric highlights could be caused by the lens flare. We believe that using a different BRDF model to account for these effects can improve the performance on these materials. We are happy to consider this as a future direction.

\section{Self-captured images outside of the laboratory}
\begin{figure}
\centering
\includegraphics[trim={0 1cm 0 0}, clip,width=0.24\textwidth]{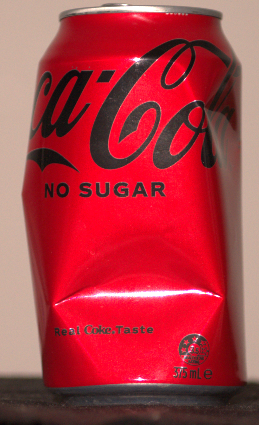}
\includegraphics[width=0.212\textwidth]{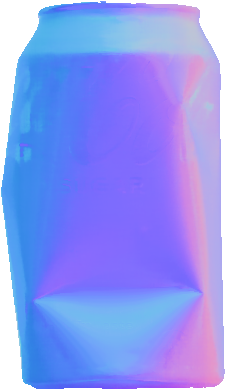}
\includegraphics[width=0.212\textwidth]{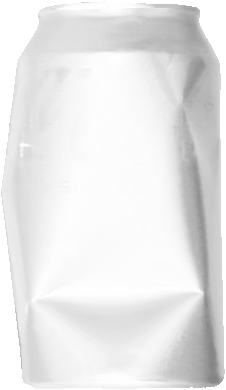}
\includegraphics[width=0.24\textwidth]{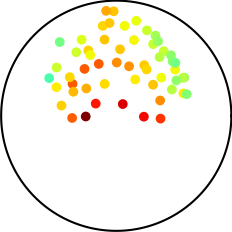}
\caption{The captured image of ``CokeCan'' and our estimations.
}
\label{fig:real}
\end{figure}

We captured $55$ images with a Nikon camera and a handheld flashlight. The target object is captured in a regular livingroom environment with lights off. 
The captured image and our estimated results (normals, shadings, and lights) are shown in \cref{fig:real}. As we can see from the results, our method still performs very well in a non-laboratory environment.

\section{Future works}
We believe that our method, with some adaptations, can be extended to solve the problem under many other assumptions, such as specularity detection, multi-view photometric stereo, photometric stereo under multi-light-sources and natural illumination.
Our method inverse renders the object to shapes and materials. Hence, the specularity detection is also available at output, as shown in \cref{fig:cat_diff_spec}.
A possible adaptation for multi-view photometric stereo is to apply our algorithm to each view of the object, and then fuse the normal map from different views to obtain the full geometry. 
We can also model the environment map as Spherical-Gaussians to enable fast integration of BRDF and lighting in natural-illumination and multi-light-sources.

\end{document}